\documentclass{article} 
\usepackage{SimpleGuideSkipConn,times}
\usepackage{hyperref}
\usepackage{url}

\usepackage{caption}
\usepackage{subcaption}
\usepackage{tikz}
\usepackage{amsmath}
\usepackage{graphicx}
\usepackage{listings}
\usepackage{csquotes}
\usepackage{xcolor}
\usepackage[colorinlistoftodos]{todonotes}
\usepackage{natbib}

\newcommand{\etal}{\emph{et al. }}

\title{A disciplined approach to neural network hyper-parameters: Part 1 -- learning rate, batch size, momentum, and weight decay}

\author{Leslie N. Smith \\
US Naval Research Laboratory\\
Washington, DC, USA \\
\texttt{leslie.smith@nrl.navy.mil} \\
}

\iclrfinalcopy 

\begin{document}

\maketitle

\begin{abstract}
Although deep learning has produced dazzling successes for applications of image, speech, and video processing in the past few years, most trainings are with suboptimal hyper-parameters, requiring unnecessarily long training times.  Setting the hyper-parameters remains a black art that requires years of experience to acquire.  This report proposes several efficient ways to set the hyper-parameters that significantly reduce training time and improves performance.  Specifically, this report shows how to examine the  training validation/test loss function for subtle clues of underfitting and overfitting and suggests guidelines for moving toward the optimal balance point.  Then it discusses how to increase/decrease the learning rate/momentum to speed up training.  Our experiments show that it is crucial to balance every manner of regularization for each dataset and architecture.  Weight decay is used as a sample regularizer to show how its optimal value is tightly coupled with the learning rates and momentum. Files to help replicate the results reported here are available at https://github.com/lnsmith54/hyperParam1.
\end{abstract}

\section{Introduction}
\label{sec:introduction}

The rise of deep learning (DL) has the potential to transform our future as a human race even more than it already has and perhaps more than any other technology.    Deep learning has already created significant improvements in computer vision, speech recognition, and natural language processing, which has led to deep learning based commercial products being ubiquitous in our society and in our lives.  

In spite of this success, the application of deep neural networks remains a black art, often requiring years of experience to effectively choose optimal hyper-parameters, regularization, and network architecture, which are all tightly coupled. Currently the process of setting the hyper-parameters, including designing the network architecture, requires expertise and extensive trial and error and is based more on serendipity than science. On the other hand, there is a recognized need to make the application of deep learning as easy as possible.  

Currently there are no simple and easy ways to  set hyper-parameters -- specifically, learning rate, batch size, momentum, and weight decay. A grid search or random search \citep{bergstra2012random} of the hyper-parameter space is computationally  expensive and time consuming.  Yet training time and final performance is highly dependent on good choices.  In addition, practitioners often choose one of the standard architectures (such as  residual networks \citep{he2016deep}) and the hyper-parameter files that are freely available in a deep learning framework's ``model zoo'' or from \url{github.com} but these are often sub-optimal for the practitioner's data.  

This report proposes several methodologies for finding optimal settings for several hyper-parameters.  A comprehensive approach of all hyper-parameters is valuable due to the interdependence of all of these factors.  Part 1 of this report examines learning rate, batch size, momentum, and weight decay and Part 2 will examine the architecture, regularization, dataset and task.  The goal is to provide the practitioner with practical advice that saves time and effort, yet improves performance.

The basis of this approach is based on the well-known concept of the balance between underfitting versus overfitting.  Specifically, it consists of examining the training's test/validation loss for clues of underfitting and overfitting in order to strive for the optimal set of hyper-parameters (this report uses ``test loss'' or ``validation loss'' interchangeably but both refer to use of validation data to find the error or accuracy produced by the network during training).  This report also suggests paying close attention to these clues while using cyclical learning rates \citep{smith2017cyclical} and cyclical momentum.  The experiments discussed herein indicate that the learning rate, momentum, and regularization are tightly coupled and optimal values must be determined together. 

Since this report is long, the reader who only wants the  highlights of this report can: (1) look at every Figure and caption, (2) read the paragraphs that start with \textbf{Remark}, and (2) review the hyper-parameter checklist at the beginning of Section \ref{sec:other}.


\section{Related work}
\label{sec:related}

The topics discussed in this report are related to a great deal of the deep learning literature.  See \cite{goodfellow2016deep} for an introductory text on the field.  Perhaps most related to this work is the book ``Neural networks: tricks of the trade'' \citep{orr2003neural} that contains several chapters with practical advice on hyper-parameters, such as \cite{bengio2012practical}.  Hence, this section only discusses a few of the most relevant papers.

This work builds on earlier work by the author.  In particular, cyclical learning rates were introduced by \cite{smith2015no} and later updated in \cite{smith2017cyclical}.  Section \ref{sec:CLR} provides updated experiments on super-convergence \citep{smith2017super}.  There is a discussion in the literature on modifying  the batch size instead of the learning rate, such as discussed in \cite{smith2017don}.  


Several recent papers discuss the use of large learning rate and small batch size, such as \cite{jastrzebski2017residual,jastrzkebski2017three,xing2018walk}.  They demonstrate that the ratio of the learning rate over the batch size guides training.  The recommendations in this report differs from those papers on the optimal setting of learning rates and batch sizes.

Smith and Le \citep{smith2017understanding} explore batch sizes and correlate the optimal batch size to the learning rate, size of the dataset, and momentum.  This report is  more comprehensive and more practical in its focus.  In addition,  Section \ref{sec:TBS} recommends a larger batch size than this paper.  

A recent paper questions the use of regularization by weight decay and dropout \citep{hernandez2018deep}.  One of the findings of this report is that the total regularization needs to be in balance for a given dataset and architecture.  Our experiments suggest that their perspective on regularization is limited -- they only add regularization by data augmentation to replace the regularization by weight decay and dropout without a full study of regularization.

There also exist approaches to learn optimal hyper-parameters by differentiating the gradient with respect to the hyper-parameters (for example see \cite{lorraine2018stochastic}).  The approach in this report is simpler for the practitioner to perform.


\begin{figure} [tbh]
	\centering
	\begin{subfigure}[b]{0.47\textwidth}
		\includegraphics[width=\textwidth]{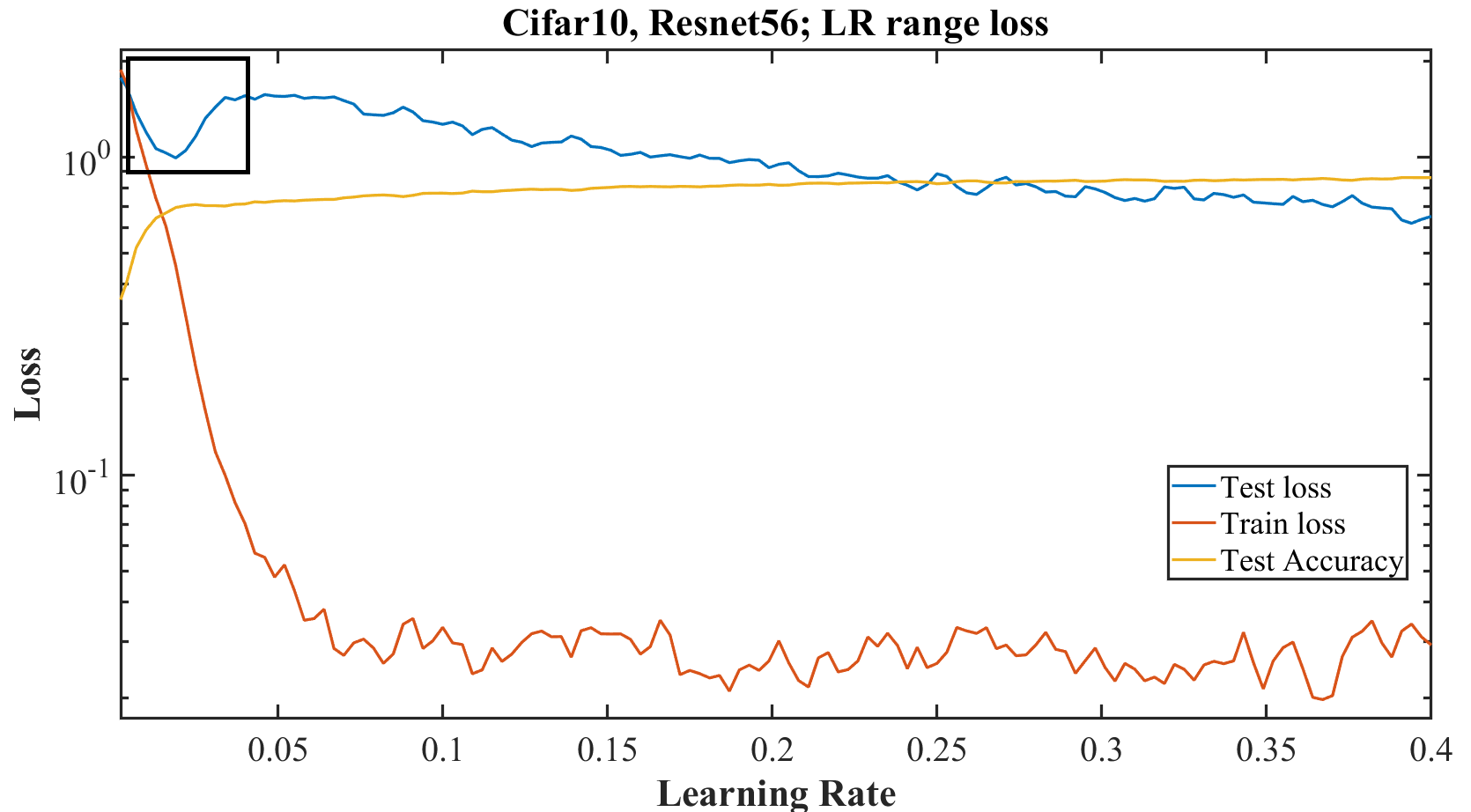}
		\caption{Characteristic plot of training loss, validation accuracy, and validation loss.  }
		\label{fig:testLoss1}       
	\end{subfigure}
	\quad
	\hfill
	~ 
	\centering
	\begin{subfigure}[b]{0.47\textwidth}
		\includegraphics[width=\textwidth]{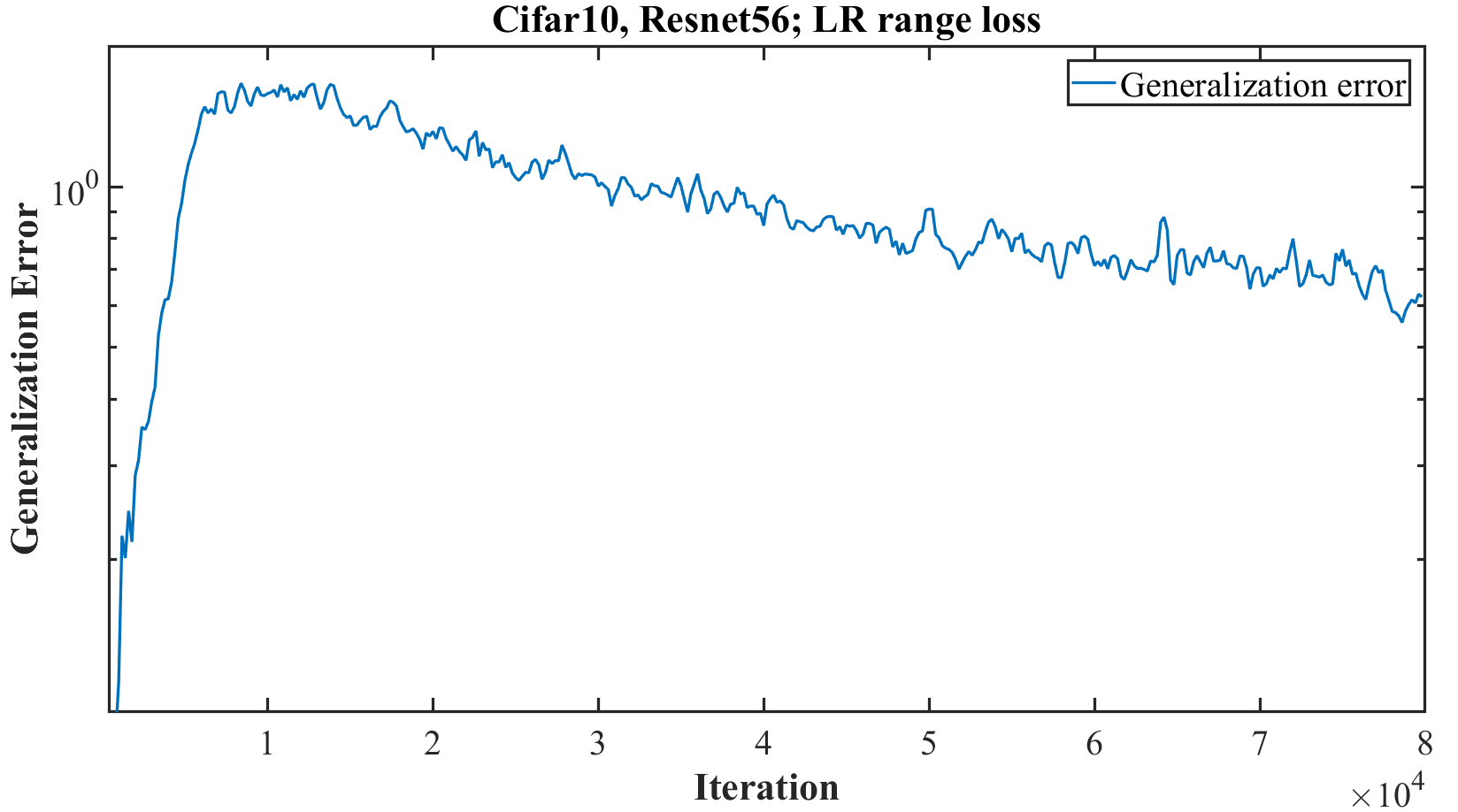}
		\caption{Characteristic plot of the generalization error, which is the validation/test loss minus the training loss. }
		\label{fig:generalizationError}       
	\end{subfigure}
	\caption{Comparison of the training loss, validation accuracy, validation loss, and generalization error that illustrates the additional information about the training process in the test/validation loss but is not visible in the test accuracy and training loss or clear with  the generalization error. These runs are a learning rate range test with the resnet-56 architecture and Cifar-10 dataset.}
	\label{fig:testLossClues}
	\vspace{-5pt}	
\end{figure}

\section{The unreasonable effectiveness of validation/test loss}
\label{sec:method}

\begin{displayquote}
``Well begun is half done.'' Aristotle
\end{displayquote}

A good detective observes subtle clues that the less observant miss.  The purpose of this Section is to draw your attention to the clues in the training process and provide guidance as to their meaning.   Often overlooked elements from the training process tell a story. By observing and understanding the clues available early during training, we can tune our architecture and hyper-parameters with short runs of a few epochs (an epoch is defined as once through the entire training data).  In particular, by monitoring validation/test loss  early in the training, enough information is available to tune the architecture and hyper-parameters and this eliminates the necessity of running complete grid or random searches.  

Figure \ref{fig:testLoss1} shows plots of the training loss, validation accuracy, and validation loss for a learning rate range test of a residual network on the Cifar dataset to find reasonable learning rates for training.  In this situation, the test loss within the black box indicates signs of overfitting at learning rates of $0.01 - 0.04$.  This information is not present in the test accuracy or in the training loss curves.  However, if we were to subtract the training loss from the test/validation loss (i.e., the generalization error) the information is present in the generalization error but often the generalization error is less clear than the validation loss. This is an example where the test loss provides valuable information.  We know that this architecture has the capacity to overfit and that early in the training too small a learning rate will create  overfitting.  

\textbf{Remark 1.} \emph{The test/validation loss is a good indicator of the network's convergence} and should be examined for clues.
In this report, the test/validation loss is used to provide insights on the training process and the final test accuracy is used for comparing performance.

Section \ref{sec:fitting} starts a brief review on the underfitting and overfitting tradeoff and demonstrates that the early training test loss provides information on how to modify the hyper-parameters.

\begin{figure}[htb]
	\center
	\includegraphics[scale=.6]{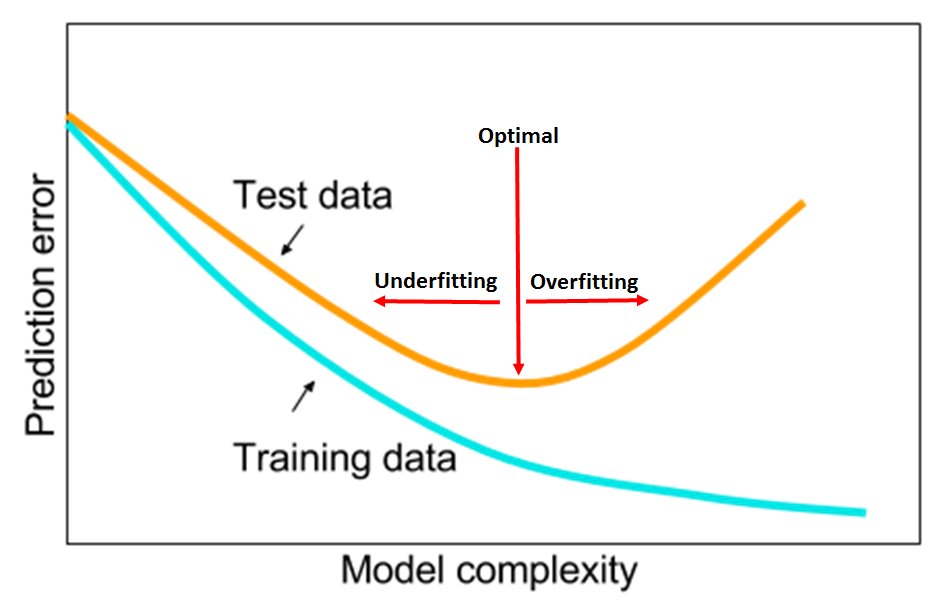}
	\caption{Pictorial explanation of the tradeoff between underfitting and overfitting.  Model complexity (the x axis) refers to the capacity or powerfulness of the machine learning model.  The figure shows the optimal capacity that falls between underfitting and overfitting.}
	\label{fig:bias-variance}       
\end{figure}

\subsection{A review of the underfitting and overfitting trade-off}
\label{sec:fitting}

Underfitting is when the machine learning model is unable to reduce the error for either the test or training set.  The cause of underfitting is an under \emph{capacity} of the machine learning model; that is, it is not powerful enough to fit the underlying complexities of the data distributions.  Overfitting happens when the machine learning model is so powerful as to fit the training set too well and the generalization error increases.  The representation of this underfitting and overfitting trade-off displayed in Figure \ref{fig:bias-variance}, which implies that achieving a horizontal test loss can point the way to the optimal balance point.  Similarly, examining the test loss during the training of a network can also point to the optimal balance of the hyper-parameters.  

\textbf{Remark 2.} The takeaway is that \emph{achieving the horizontal part of the test loss is the goal of hyper-parameter tuning.}
Achieving this balance can be difficult with deep neural networks.   Deep networks are very powerful, with networks becoming more powerful with greater depth (i.e., more layers), width (i.e, more neurons or filters per layer), and the addition of  skip connections to the architecture.   Also, there are various forms of regulation, such as weight decay or dropout \citep{srivastava2014dropout}.  One needs to vary important hyper-parameters and can use a variety of optimization methods, such as Nesterov or Adam \citep{kingma2014adam}.  It is well known that optimizing all of these elements to achieve the best performance on a given dataset is a challenge.  




An insight that inspired this Section is that signs of underfitting or overfitting of the test or validation loss early in the training process are useful for tuning the hyper-parameters.  This section started with the quote ``Well begun is half done'' because substantial time can be saved by attending to the test loss early in the training.  For example, Figure \ref{fig:testLoss1} shows some overfitting within the black square that indicates a sub-optimal choice of hyper-parameters.  If the hyper-parameters are set well at the beginning, they will perform well through the entire training process.   In addition, if the hyper-parameters are set using only a few epochs, a significant time savings is possible in the search for hyper-parameters.  The test loss during the training process can be used to find the optimal network architecture and hyper-parameters without performing a full training in order to compare the final performance results.

The rest of this report discusses the early signs of underfitting and overfitting that are visible in the test loss.  In addition, it discusses how adjustments to the hyper-parameters affects underfitting and overfitting.  This is necessary in order to know how to adjust the hyper-parameters.  

\begin{figure} [tbh]
	\centering
	\begin{subfigure}[b]{0.4\textwidth}
		\includegraphics[width=\textwidth]{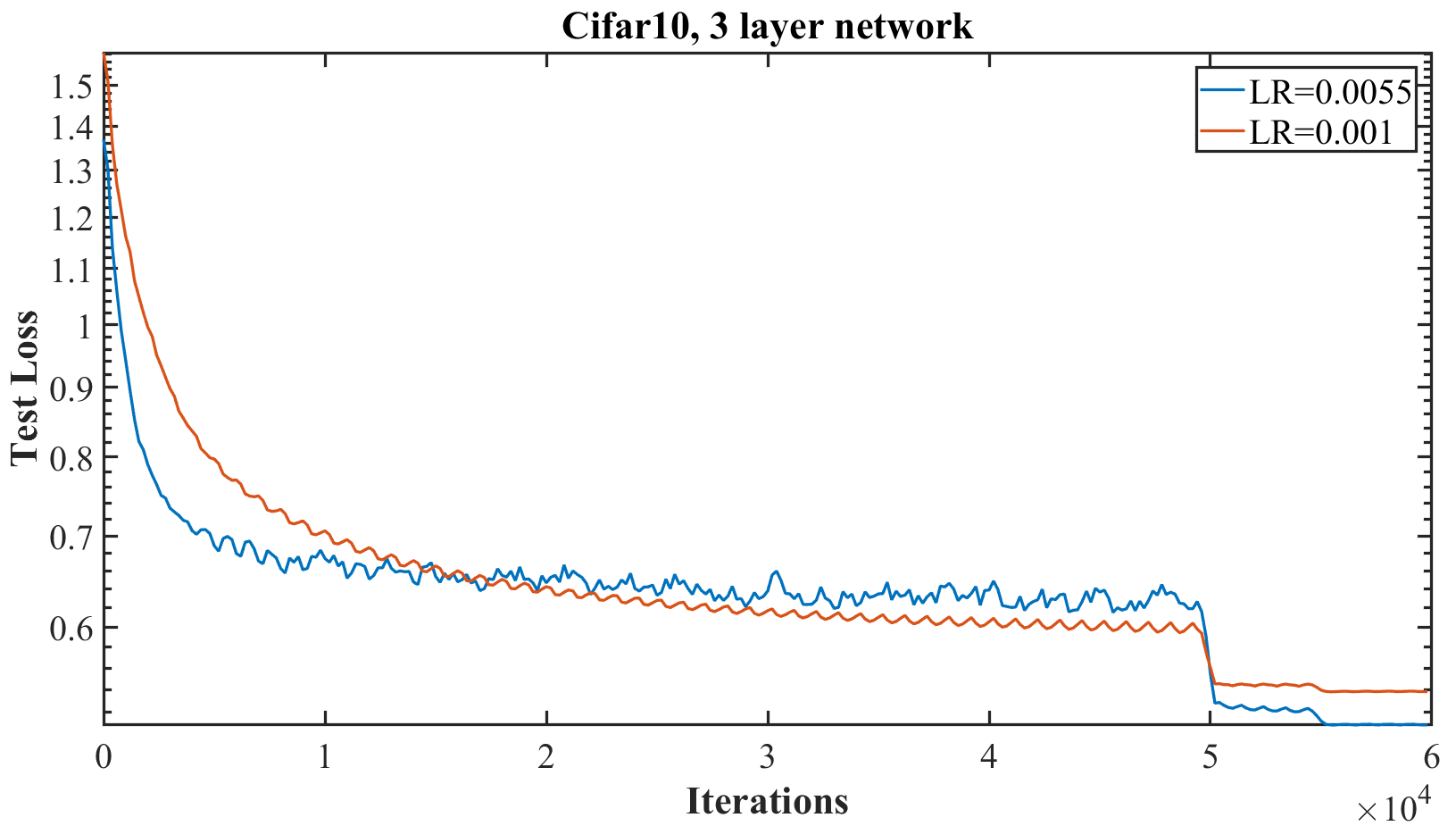}
		\caption{Test loss for the Cifar-10 dataset with a shallow 3 layer network. }
		\label{fig:3layerLoss}       
	\end{subfigure}
	\quad
	\hfill
	~ 
	\centering
	\begin{subfigure}[b]{0.52\textwidth}
		\includegraphics[width=\textwidth]{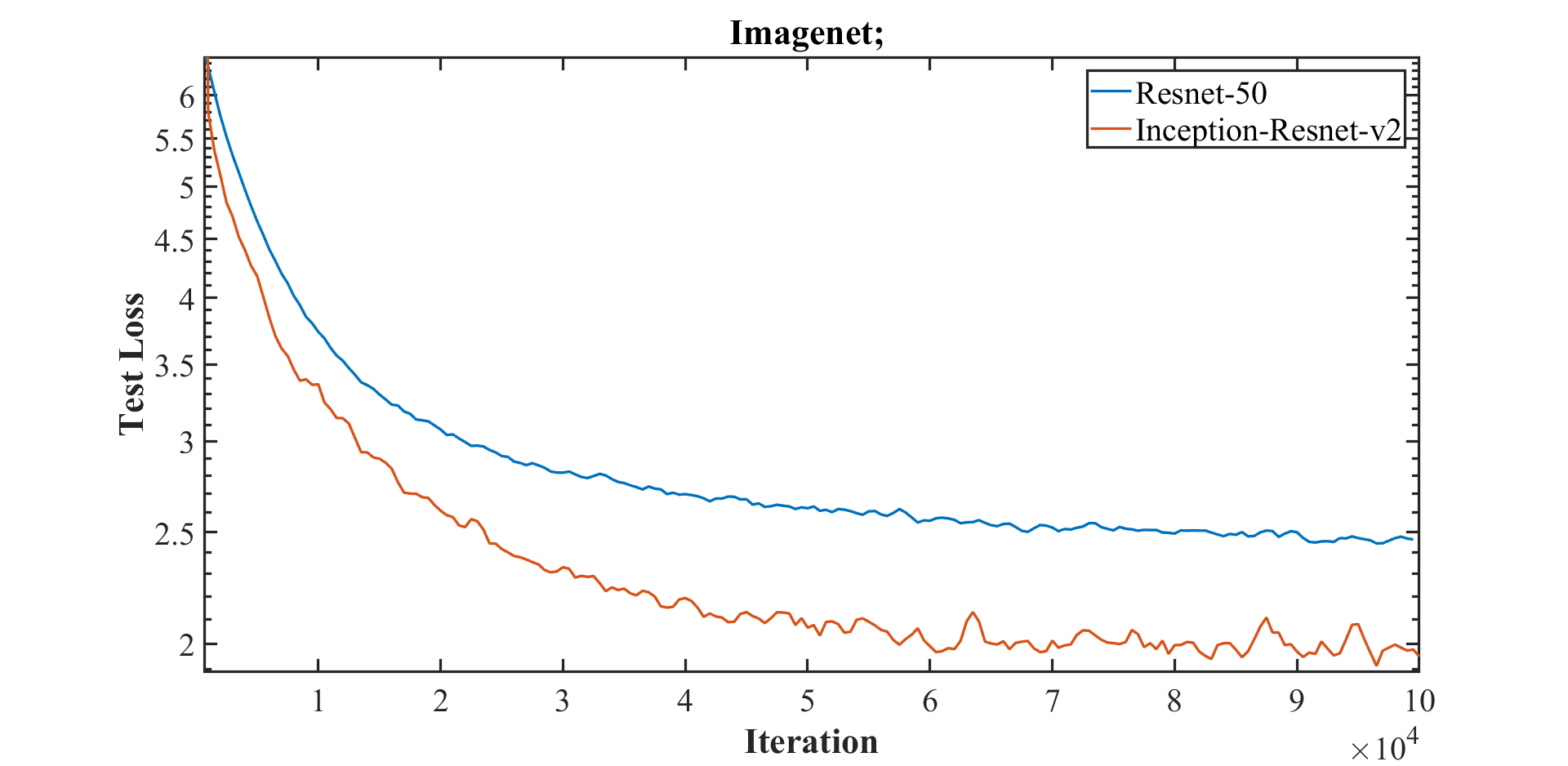}
		\caption{Test loss for Imagenet with two networks; resnet-50 and inception-resnet-v2.}
		\label{fig:imagenetTestLoss3}       
	\end{subfigure}
	\caption{Underfitting is characterized by a continuously decreasing test loss, rather than a horizontal plateau. Underfitting is  visible during the training on two different datasets, Cifar-10 and imagenet.}
	\label{fig:Loss1}
	\vspace{-5pt}	
\end{figure}

\subsection{Underfitting}
\label{sec:underfitting}

Our first example is with a shallow, 3-layer network on the Cifar-10 dataset.  The red curve in Figure \ref{fig:3layerLoss} with a learning rate of 0.001 shows a decreasing test loss.  This  curve indicates underfitting because it continues to decrease, like the left side of the test loss curve in Figure \ref{fig:bias-variance}. Increasing the learning rate moves the training from underfitting towards overfitting.  The blue curve shows the test loss with a learning rate of 0.004.  Note that the test loss decreases more rapidly during the initial iterations and is then horizontal.  This is one of the early positive clues that indicates that this curve's configuration will produce a better final accuracy than the other configuration, which it does.  

The second example is on the Imagenet dataset with two architectures: resnet-50, and inception-resnet-v2.  Here the cause of underfitting is due to the underlying complexities of the data distributions.  In Figure \ref{fig:imagenetTestLoss3} the test loss continues to decrease over the 100,000 iterations (about 3 epochs) but the inception-resnet-v2 decreases more and becomes more horizontal, indicating that the inception-resnet-v2 has less underfitting.

Increasing the learning rate helps reduce underfitting.  An easy way to find a good learning rate is the LR range test \citep{smith2017cyclical} (additional description of the LR range test is in Section \ref{sec:CLR}).  

\begin{figure} [tbh]
	\centering
	\begin{subfigure}[b]{0.47\textwidth}
		\includegraphics[width=\textwidth]{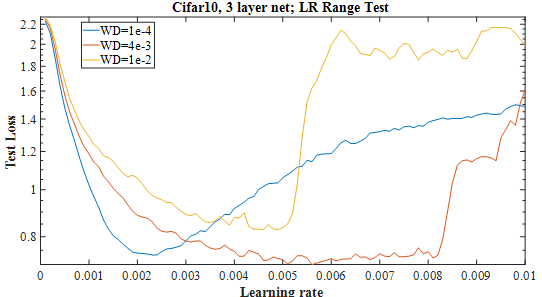}  
		\caption{Cifar-10 dataset with a shallow 3 layer network.}
		\label{fig:overfitting3}       
	\end{subfigure}
	\quad
	\hfill
	~ 
	\centering
	\begin{subfigure}[b]{0.47\textwidth}
		\includegraphics[width=\textwidth]{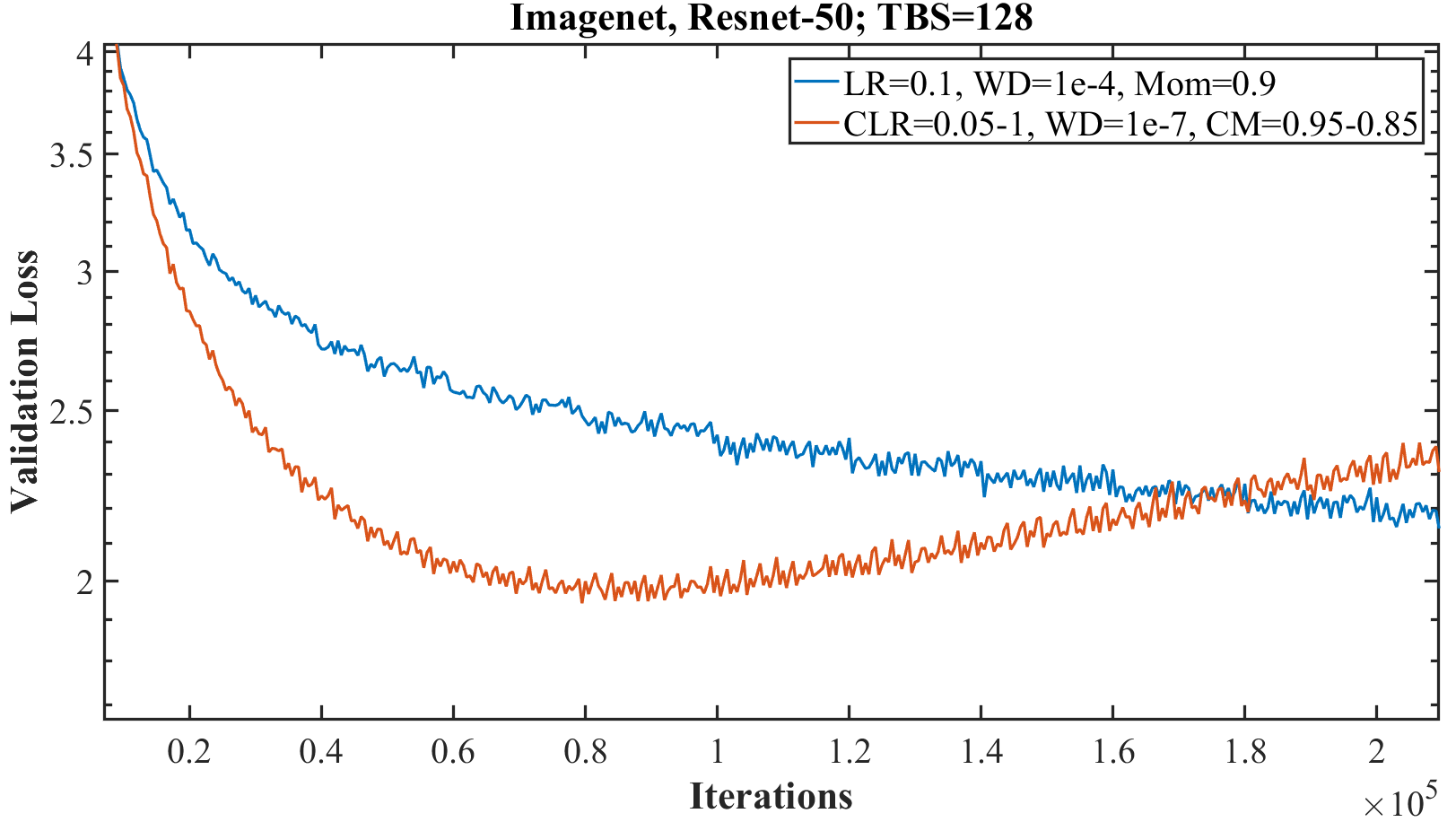}
		\caption{Imagenet dataset with resnet-50 architecture.}
		\label{fig:imagenetResnetOverfitting}       
	\end{subfigure}
	\caption{Increasing validation/test loss indicates overfitting.  Examples of overfitting are shown for Cifar-10 and Imagenet. WD = weight decay, LR = learning rate, CLR = cyclical learning rate, CM = cyclical momentum.}
	\label{fig:OverfittingEx}
	\vspace{-15pt}	
\end{figure}

\subsection{Overfitting}
\label{sec:overfitting}

Overfitting tends to be more complicated than underfitting but clues are visible in the test loss.  In Figure \ref{fig:bias-variance} the test loss goes from underfitting (decreasing) to overfitting (increasing) but overfitting in neural networks is often not so simple.  In Figure \ref{fig:testLossClues} there are signs of overfitting in the test loss at small learning rates (0.01 - 0.04) but then the test loss continues to decrease at higher learning rates as though it is underfitting.  This indicates that learning rates that are too small can exhibit some overfitting behavior.

Figure \ref{fig:overfitting3} displays test loss for training a shallow, 3-layer network trained on Cifar-10.  The figure shows the results of a learning rate range test at three values for weight decay (WD).  At WD = $10^{-4}$ the loss reaches a minimum loss near a learning rate of 0.002, then begins to increase, displaying overfitting.  If WD = $4 \times 10^{-3}$, the loss is stable over a larger range of learning rate values and the loss attains a lower loss value, indicating that the latter WD value is superior to the former.  Also note that the sharp increase in loss for the yellow curve near the learning rate of 0.005 is not a sign of overfitting but is caused by instabilities in the training due to the large learning rate.  Similarly, the red curve diverges above a learning rate of 0.008.

The next example is training the Imagenet dataset with a resnet-50 architecture.  While the blue curve in Figure \ref{fig:imagenetResnetOverfitting}  shows underfitting with weight decay of $10^{-4}$ and a constant learning rate of 0.1, the red curve shows overfitting with too small a weight decay of $10^{-7}$.  Both curves indicate non-optimal settings of the hyper-parameters.  

An additional example of overfitting is visible with the yellow curve in Figure \ref{fig:3layerMomCLRTestLoss} as it slowly rises above LR=0.006 and until divergence near LR=0.01.  The blue curves in Figure \ref{fig:3layerWDTestLoss} and in Figure \ref{fig:ResnetCifarWDLRtestLoss} show two additional forms of overfitting.  The former illustrates a more gradual increase in the error while the later shows a temporary initial increase (additional analysis of this Figure is given in Section \ref{sec:imagenet}). All situations are examples of incorrectly set hyper-parameters.

The art of setting the network's hyper-parameters amounts to ending up at the balance point between underfitting and overfitting.

\section{Cyclical learning rates, batch sizes, cyclical momentum, and weight decay}
\label{sec:cyclicalMomentum}

Choosing learning rate, momentum, and weight decay hyper-parameters well will improve the network's performance.  The conventional method is to perform a grid or a random search, which can be computationally expensive and time consuming. In addition, the effects of these hyper-parameters are tightly coupled with each other, the data, and architecture. This section offers more efficient ways to choose these hyper-parameters.

\subsection{Cyclical learning rates and super-convergence revisited}
\label{sec:CLR}

If the learning rate (LR) is too small, overfitting can occur.  Large learning rates help to regularize the training but if the learning rate is too large, the training will diverge.  Hence a grid search of short runs to find learning rates that converge or diverge is possible but there is an easier way.

Cyclical learning rates (CLR) and the learning rate range test (LR range test) were first proposed by \cite{smith2015no} and later updated in \cite{smith2017cyclical} as a recipe for choosing the learning rate.  Here is a brief review; refer to the  original papers for more details.

To use CLR, one specifies minimum and maximum learning rate boundaries and a stepsize.  The stepsize is the number of iterations (or epochs) used for each step and a cycle consists of two such steps -- one in which the learning rate linearly increases from the minimum to the maximum and the other in which it linearly decreases.  \cite{smith2015no} tested numerous ways to vary the learning rate between the two boundary values, found them to be equivalent  and therefore recommended the simplest, which is letting the learning rate change linearly (\cite{jastrzkebski2017three} suggest discrete jumps and obtained similar results).  

In the LR range test, training starts with a small learning rate which is slowly increased linearly throughout a  pre-training run.  This single run provides valuable information on how well the network can be trained over a range of learning rates and what is the maximum learning rate.   When starting with a small learning rate, the network begins to converge and, as the learning rate increases, it eventually becomes too large and causes the test/validation loss to increase and the accuracy to decrease. The learning rate at this extrema is the largest value that can be used as the  learning rate for the maximum bound with cyclical learning rates but a smaller value will be necessary when choosing a constant learning rate or the network will not begin to converge.  There are several ways one can choose the minimum learning rate bound: (1) a factor of 3 or 4 less than the maximum bound, (2) a factor of 10 or 20 less than the maximum bound if only one cycle is used, (3)  by a short test of hundreds of iterations with a few initial learning rates and pick the largest one that allows convergence to begin without signs of overfitting as shown in Figure \ref{fig:testLoss1} (if the initial learning rate is too large, the training won't begin to converge).  Take note that there is a maximum speed the learning rate can increase without the training becoming unstable, which effects your choices for the minimum and maximum learning rates (i.e., increase the stepsize to increase the difference between the minimum and maximum).

\begin{figure} [tbh]
	\centering
	\begin{subfigure}[b]{0.42\textwidth}
		\includegraphics[width=\textwidth]{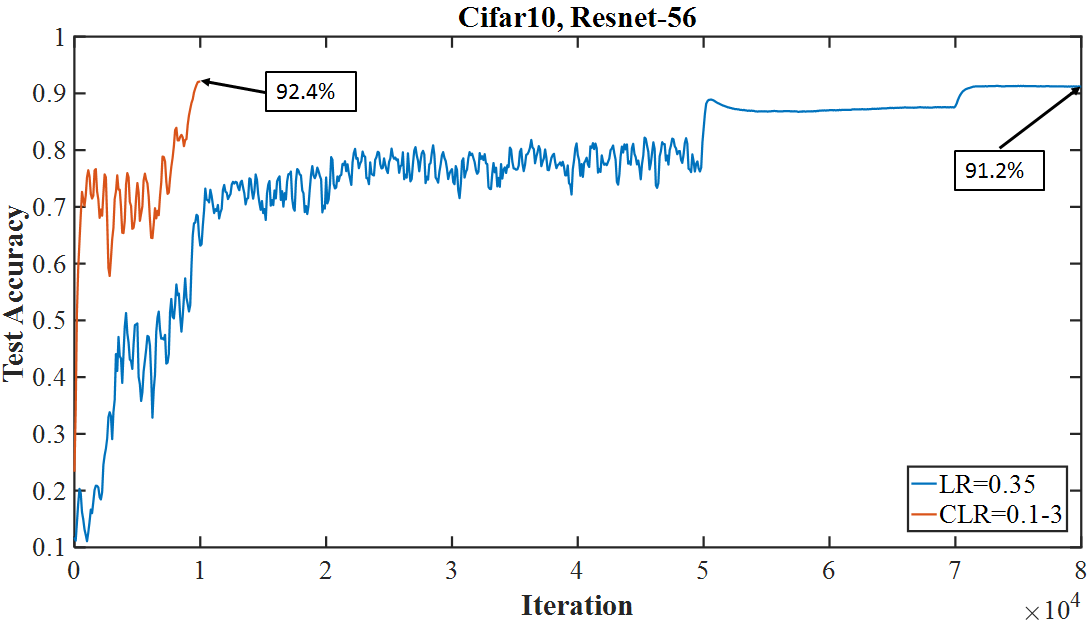}
		\caption{An example of super-convergence.}
		\label{fig:LRvsCLRresnet56}       
	\end{subfigure}
	\quad
	\hfill
	~ 
	\centering
	\begin{subfigure}[b]{0.52\textwidth}
		\includegraphics[width=\textwidth]{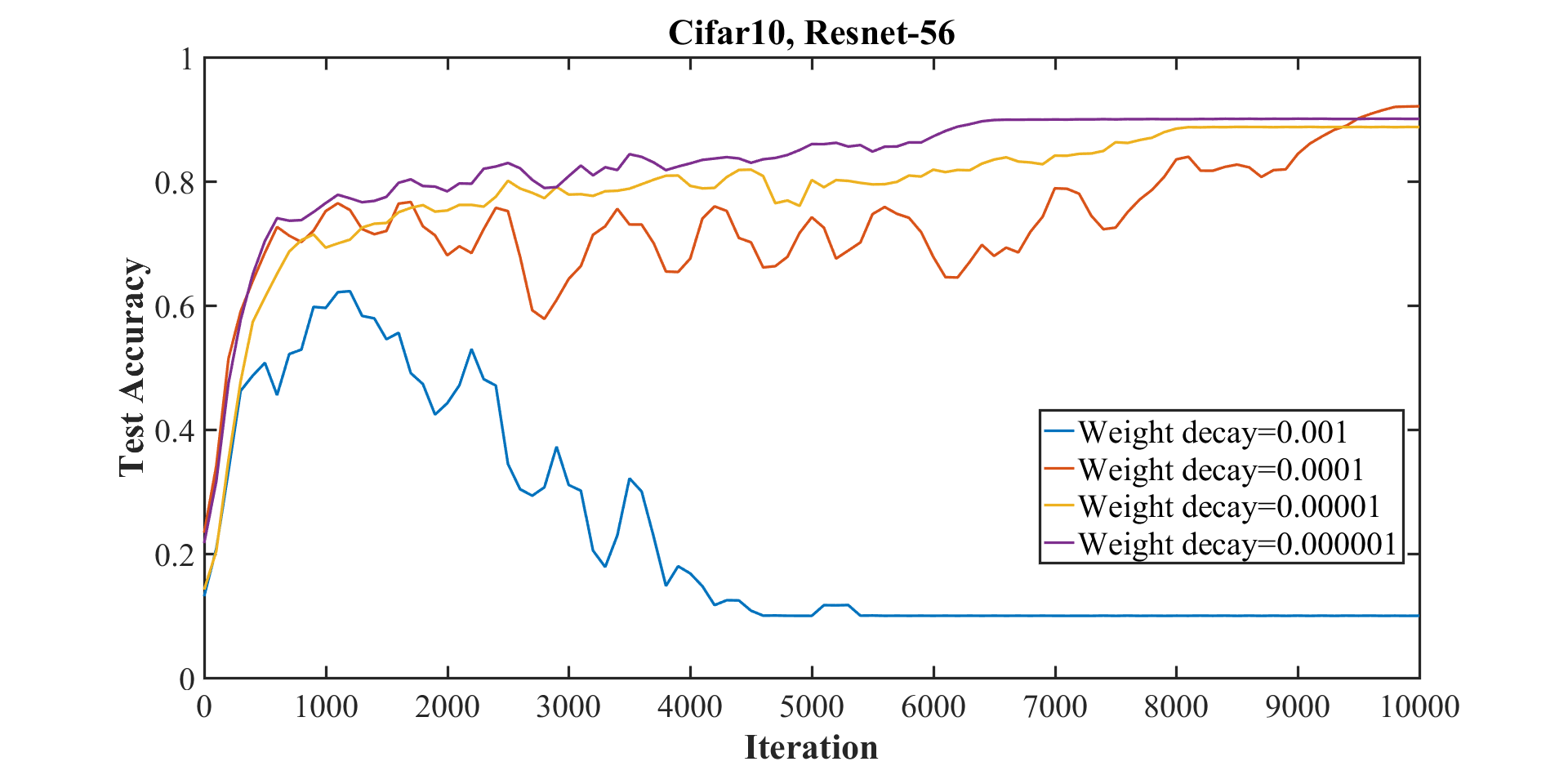}
		\caption{The effect of weight decay.}
		\label{fig:clr3SS5kResnet56WD}       
	\end{subfigure}
	\caption{Faster training is possible by allowing the learning rates to become large.  Other regularization methods must be reduced to compensate for the regularization effects of large learning rates. Practitioners must strive for an optimal balance of regularization.}
	\label{fig:superConvergence}
	\vspace{-5pt}	
\end{figure}

Super-convergence \citep{smith2017super} was shown to happen when using deep resnets on cifar10 or cifar-100 data, where the test loss and accuracy remain nearly constant for this LR range test, even up to very large learning rates.  In these situations the  network can be trained quickly with one learning rate cycle by using an unusually large learning rate. The very large learning rates used provided the twin benefits of regularization that prevented overfitting and faster training of the network.  Figure \ref{fig:LRvsCLRresnet56} shows an example of super-convergence, where the training was completed in 10,000 iterations  by using learning rates up to 3.0 instead of needing 80,000 iterations with a constant initial learning rate of 0.1.  

Here we suggest a slight modification of cyclical learning rate policy for super-convergence; always use one cycle that is smaller than the total number of iterations/epochs and allow the learning rate to decrease several orders of magnitude less than the initial learning rate for the remaining iterations.  We named this learning rate policy ``1cycle'' and in our experiments this policy allows the accuracy to plateau before the training ends.  It is interesting to note that the 1cycle learning rate policy is a combination of curriculum learning  \citep{bengio2009curriculum} and simulated annealing \citep{aarts1988simulated}, both of which have a long history of use in deep learning.

This report shows that super-convergence is universal and provides additional guidance on why, when and where this is possible. There are many forms of regularization, such as large learning rates, small batch sizes, weight decay, and dropout \citep{srivastava2014dropout}.  Practitioners must balance the various forms of regularization for each dataset and architecture in order to obtain good performance.  Figure \ref{fig:clr3SS5kResnet56WD}  shows that weight decay values of $10^{-4}$ or smaller allow the use of large learning rates (i.e., up to 3) but setting weight decay of $10^{-3}$ eliminates the ability to train the networks with such a large learning rate.  It is because the regularization needs to be balanced and it is necessary to reduce other forms of regularization in order to utilize the regularization from large learning rates and gain the other benefit - faster training.

\textbf{Remark 3.}
A general principle is: \emph{the amount of regularization must be balanced for each dataset and architecture.}  Recognition of this principle permits general use of super-convergence.  Reducing other forms of regularization and regularizing with very large learning rates makes training significantly more efficient.  

Experiments in Section \ref*{sec:other} with MNIST, Cifar10, Cifar-100, and Imagenet, and various architectures, such as shallow nets,  resnets, wide resnets, densenets, inception-resnet, show that all can be trained more quickly with large learning rates, provided other forms of regularizations are reduced to an optimal balance point.  This requires experimenting with the LR range test with a variety of regularization settings.  As an example, Section \ref*{sec:WDreg} demonstrates how to find a good value of weight decay with the LR range test.  A similar procedure can be performed with other forms of regularization (i.e., choosing the dropout ratio).

\begin{figure} [tbh]
	\centering
	\begin{subfigure}[b]{0.42\textwidth}
		\includegraphics[width=\textwidth]{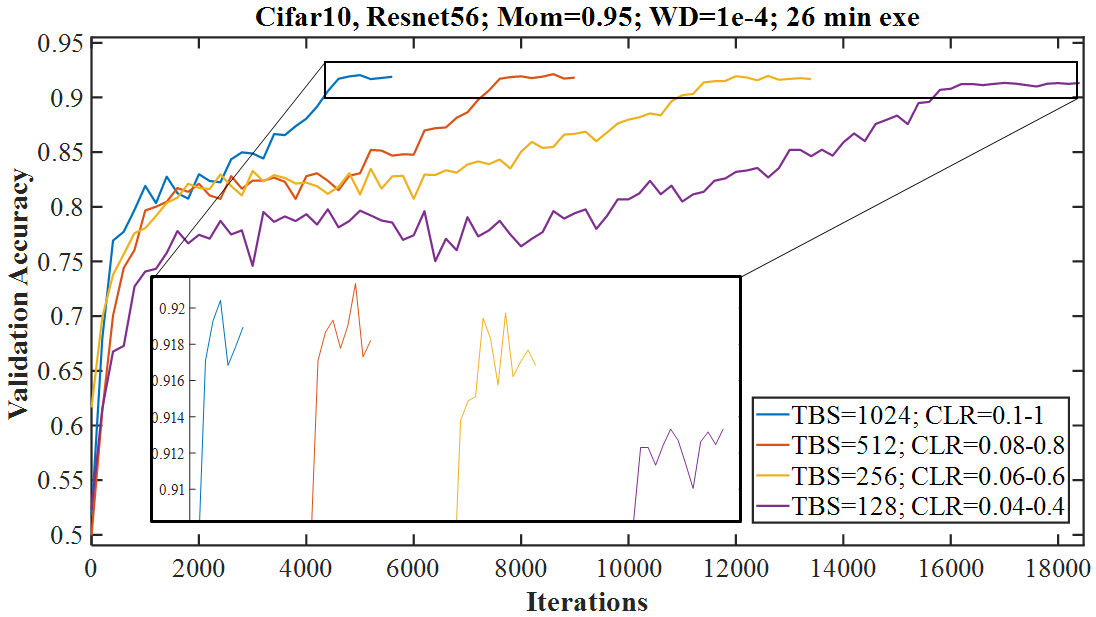}
		\caption{The effect of batch size on test accuracy.}
		\label{fig:resnet56CifarTBSAcc}       
	\end{subfigure}
	\centering
	\begin{subfigure}[b]{0.5\textwidth}
		\includegraphics[width=\textwidth]{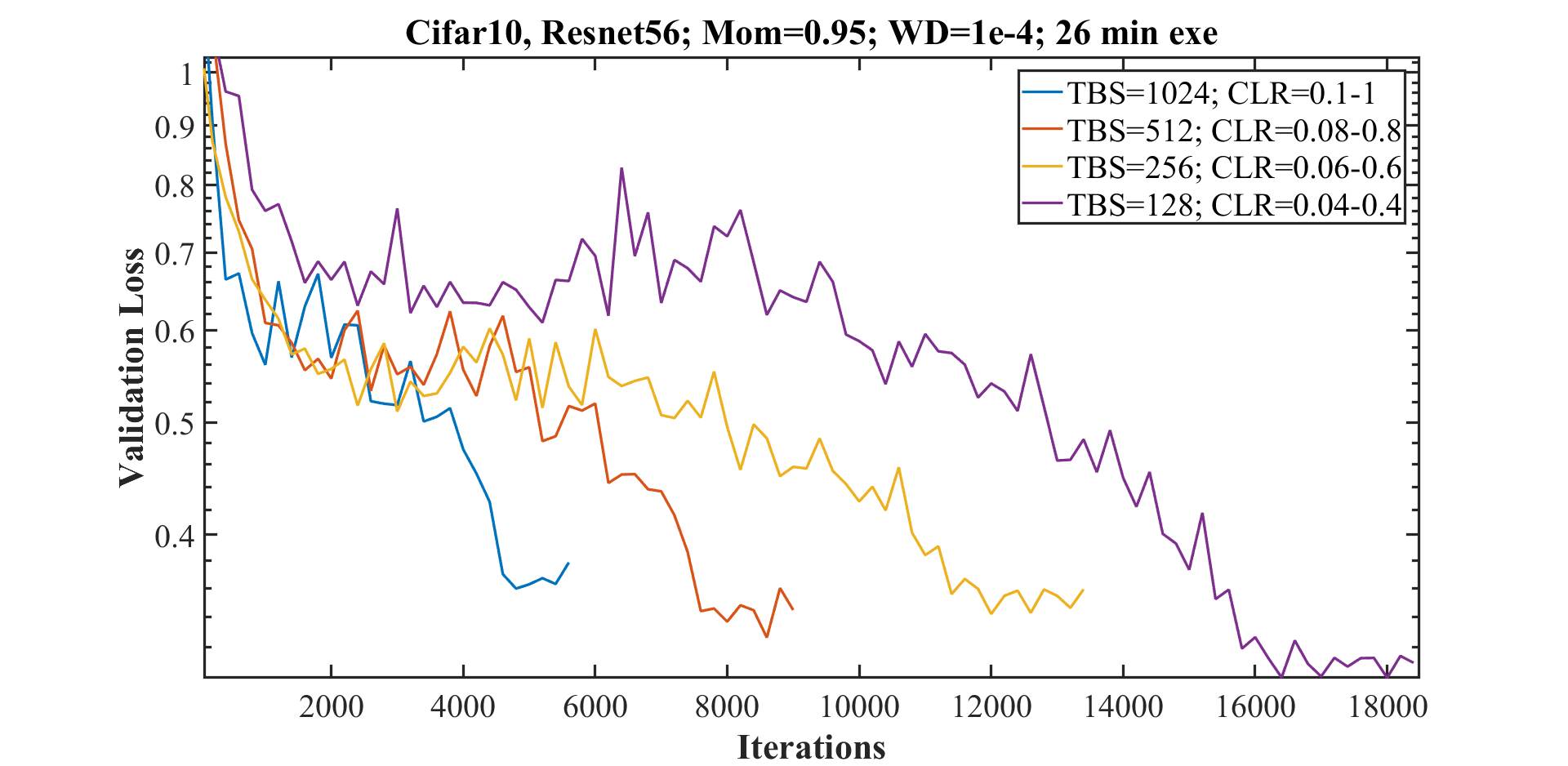}
		\caption{The effect of batch size on test loss.}
		\label{fig:resnet56CifarTBSLoss}       
	\end{subfigure}
	\caption{The effects of total batch size (TBS) on validation accuracy/loss for the Cifar-10  with resnet-56 and a 1cycle learning rate schedule.  For a fixed computational budget, larger TBS yields higher test accuracy but smaller TBS has lower test loss.}
	\label{fig:batchsize}
	\vspace{-15pt}	
\end{figure}

\subsection{Batch size}
\label{sec:TBS}

Small batch sizes have been recommended for regularization effects \citep{wilson2003general} and others have shown there to be an optimal batch size on the order of 80 for Cifar-10 \citep{smith2017understanding}. Contrary to this early work, this Section recommends using a larger batch size when using the 1cycle learning rate schedule, which is described in the above.

A difficulty in comparing batch sizes is that one obtains conflicting results if one maintains a constant number of epochs versus a constant number of iterations.  This Section suggests that neither is appropriate for comparing different batch sizes.  The constant number of epochs is inappropriate as it doesn't account for the significant computational efficiencies of larger batch sizes so it penalizes larger batch sizes.  On the other hand, a constant number of iterations favors larger batch sizes too much.  Another factor is that larger batch sizes permit the use of larger learning rates in the 1cycle learning rate schedule.   Instead our experiments aimed to compare batch sizes by maintaining a near constant execution time because practitioners are interested in minimizing training time while maintaining high performance.  

\textbf{Remark 4.}
The takeaway message of this Section is that \emph{the practitioner's goal is obtaining the highest performance while minimizing the needed computational time.}  Unlike the learning rate hyper-parameter where its value doesn't affect computational time, batch size must be examined in conjunction with the execution time of the training.  Similarly, choosing the number of epochs/iterations for training should be large enough to maximize the final test performance but no larger.

Figure \ref{fig:resnet56CifarTBSAcc} shows the validation accuracy for training resnet-56 on Cifar-10 with the 1cycle learning rate schedule. Unlike the experiments in other sections, here each curve is an average of four runs that all used the same batch sizes (rather than a small range of batch sizes).  The four curves represent four different total batch sizes (TBS) of 128, 256, 512, and 1024.  The number of iterations/epochs for all of these runs was chosen to provide a near constant execution time of 26 minutes on our IBM Power8.  This value for the execution time was found by a grid search on the minimum number of epochs needed for the TBS = 128 case to obtain an optimal accuracy.  Once this was found, the number of epochs/iterations for the other TBS cases were computed such that the training was performed approximately in the same execution time.  Larger batch sizes used larger learning rates, as shown in the legend for Figure \ref{fig:resnet56CifarTBSAcc}.

It is clear in Figure \ref{fig:resnet56CifarTBSAcc} that the larger batch sizes ran in fewer iterations.  Not shown is that larger batch sizes had more epochs (TBS/epochs = 128/48, 256/70, 512/95, 1024/116).  The legend does show the learning rate range for each TBS and that larger learning rates are used with the larger batch sizes.  The blowup compares the final accuracies for the four batch sizes.  The results imply that it is beneficial to use larger batch sizes but the benefit tapers off (the final results for TBS = 1024 is quite close to the results for TBS = 512), perhaps due to the too great a reduction in number of iterations. Hence, TBS = 512 is a good choice for this dataset, architecture, and computer hardware.   

It is also interesting to contrast the test loss to the test accuracy.  Figure \ref{fig:resnet56CifarTBSLoss} shows the validation loss for the same runs shown in Figure \ref{fig:resnet56CifarTBSAcc}.  Although the larger batch sizes have lower loss values early in the training, the final loss values are lower as the batch sizes decrease, which is the opposite performance as accuracy results.  During the course of our experiments, this kind of discrepancy between the validation loss and accuracy was unusual. 
 
Some papers have recommended modifying the batch size, rather than the learning rates (i.e., \cite{smith2017don,jastrzkebski2017three}).  However, the batch size is limited by your hardware's memory, while the learning rate is not.  While some practitioners are able to run on effectively unlimited numbers of nodes and GPUs, if this is not true for you, this report recommends you use a batch size that fits in your hardware's memory and enable using larger learning rates.

\begin{figure} [tbh]
	\centering
	\begin{subfigure}[b]{0.5\textwidth}
		\includegraphics[width=\textwidth]{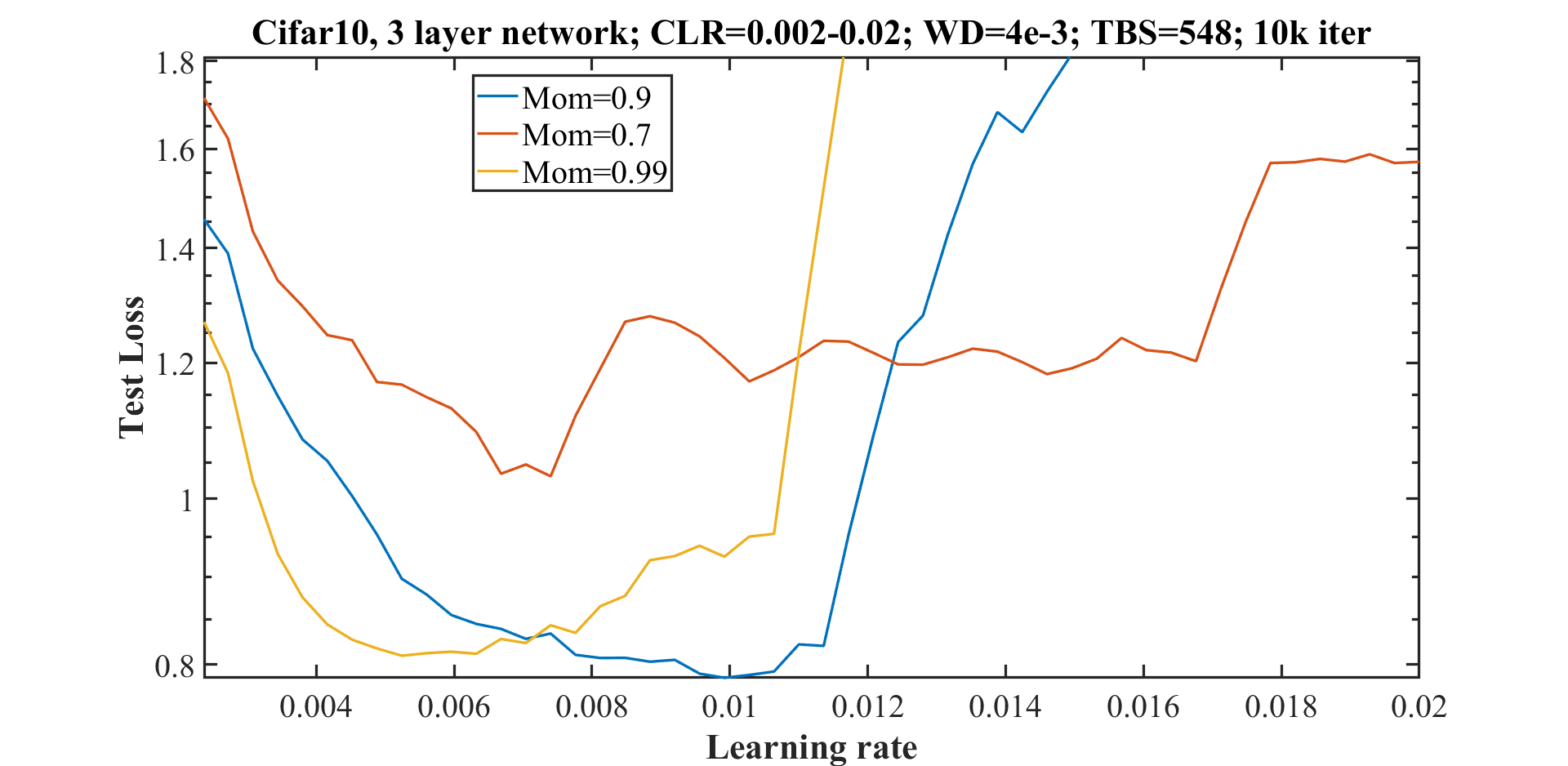}
		\caption{Showing that the value of momentum matters.}
		\label{fig:3layerMomCLRTestLoss}       
	\end{subfigure}
	\quad
	\hfill
	~ 
	\centering
	\begin{subfigure}[b]{0.43\textwidth}
		\includegraphics[width=\textwidth]{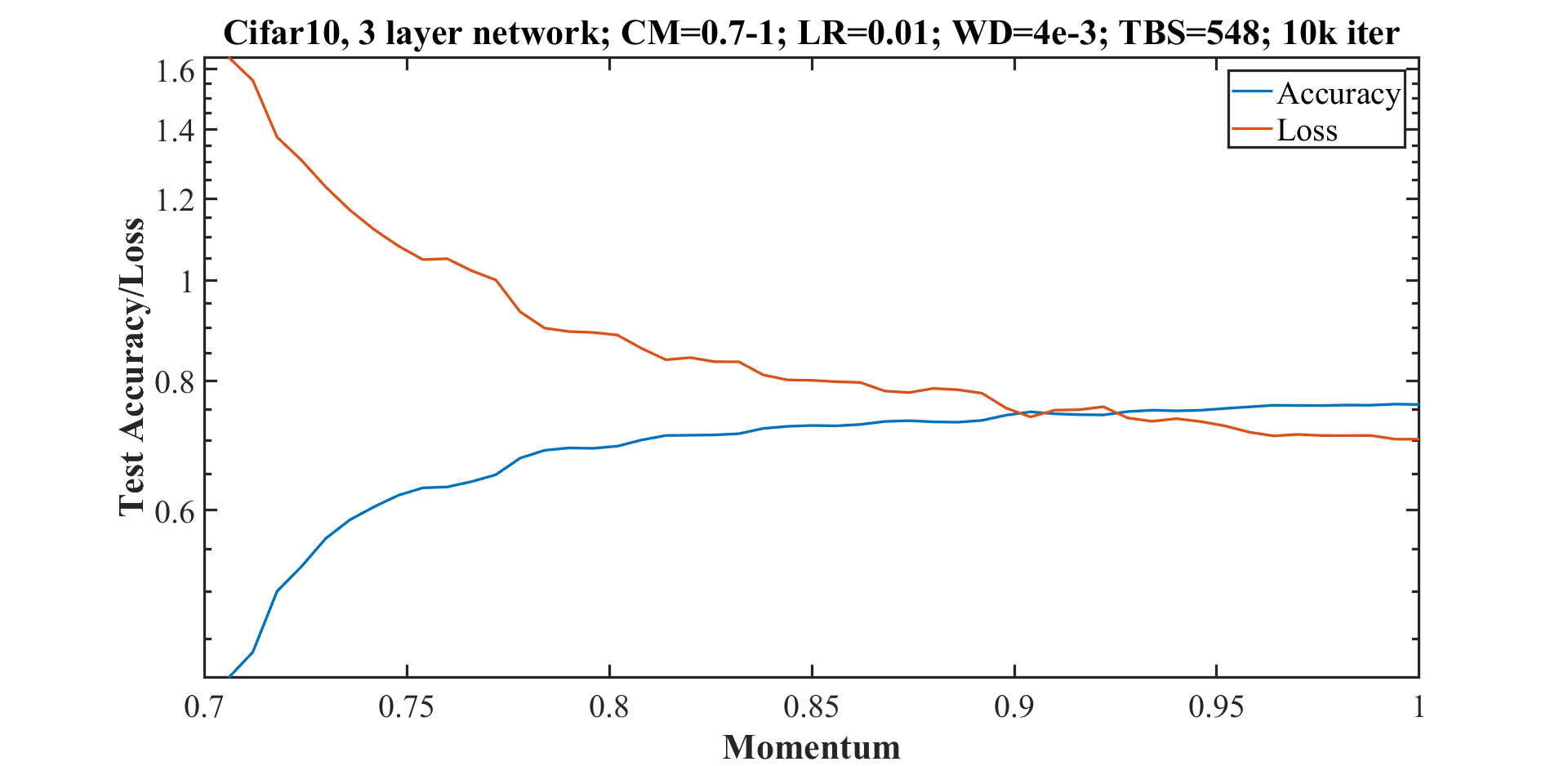}
		\caption{Increasing momentum does not find an optimal value.}
		\label{fig:3layerCMtestAccLoss}       
	\end{subfigure}
	\centering
	\begin{subfigure}[b]{0.5\textwidth}
		\includegraphics[width=\textwidth]{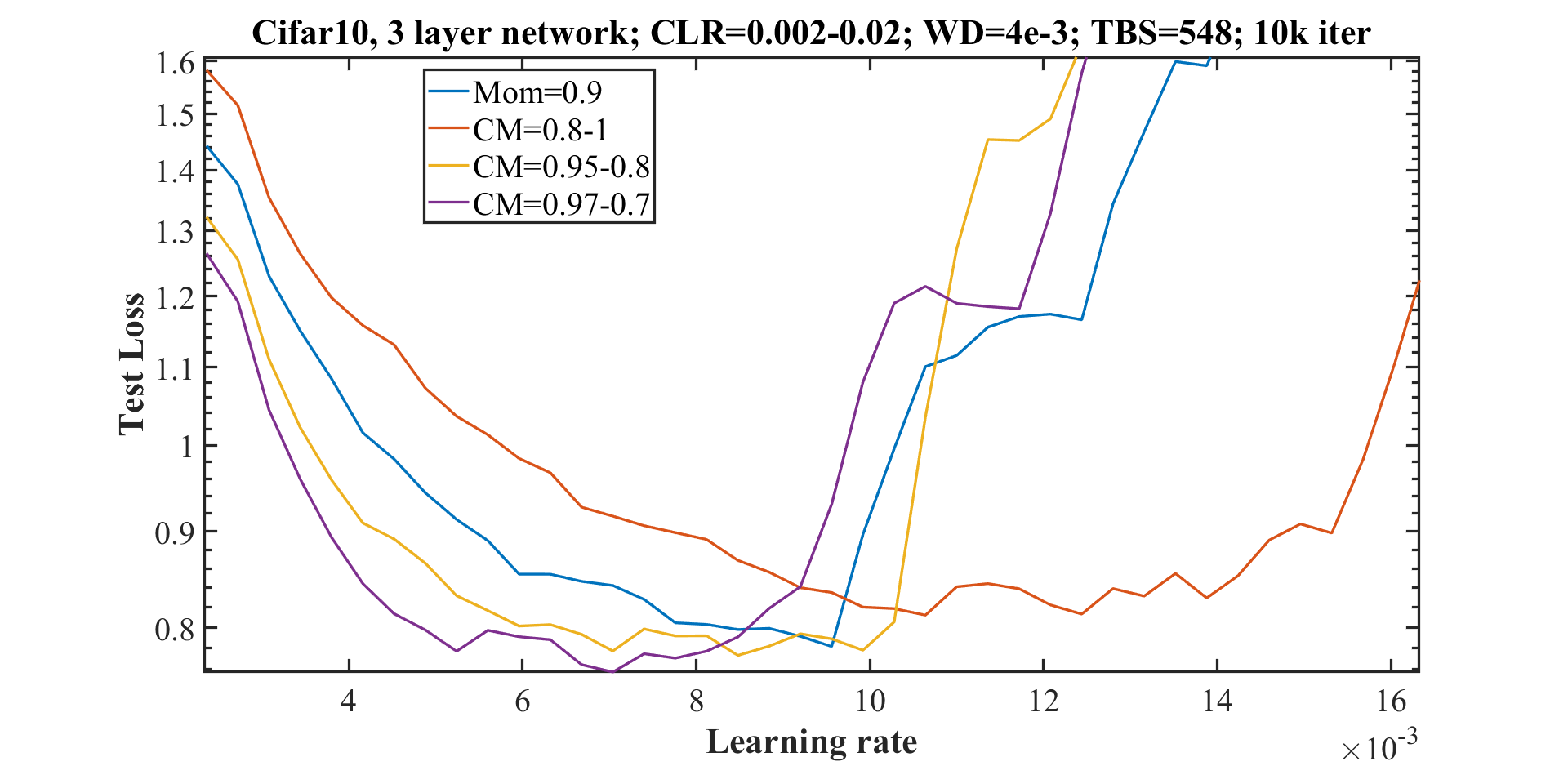}
		\caption{Cyclical momentum combined with cyclical learning rates.}
		\label{fig:3layerCMtestLoss}       
	\end{subfigure}
	\quad
	\hfill
	~ 
	\centering
	\begin{subfigure}[b]{0.43\textwidth}
		\includegraphics[width=\textwidth]{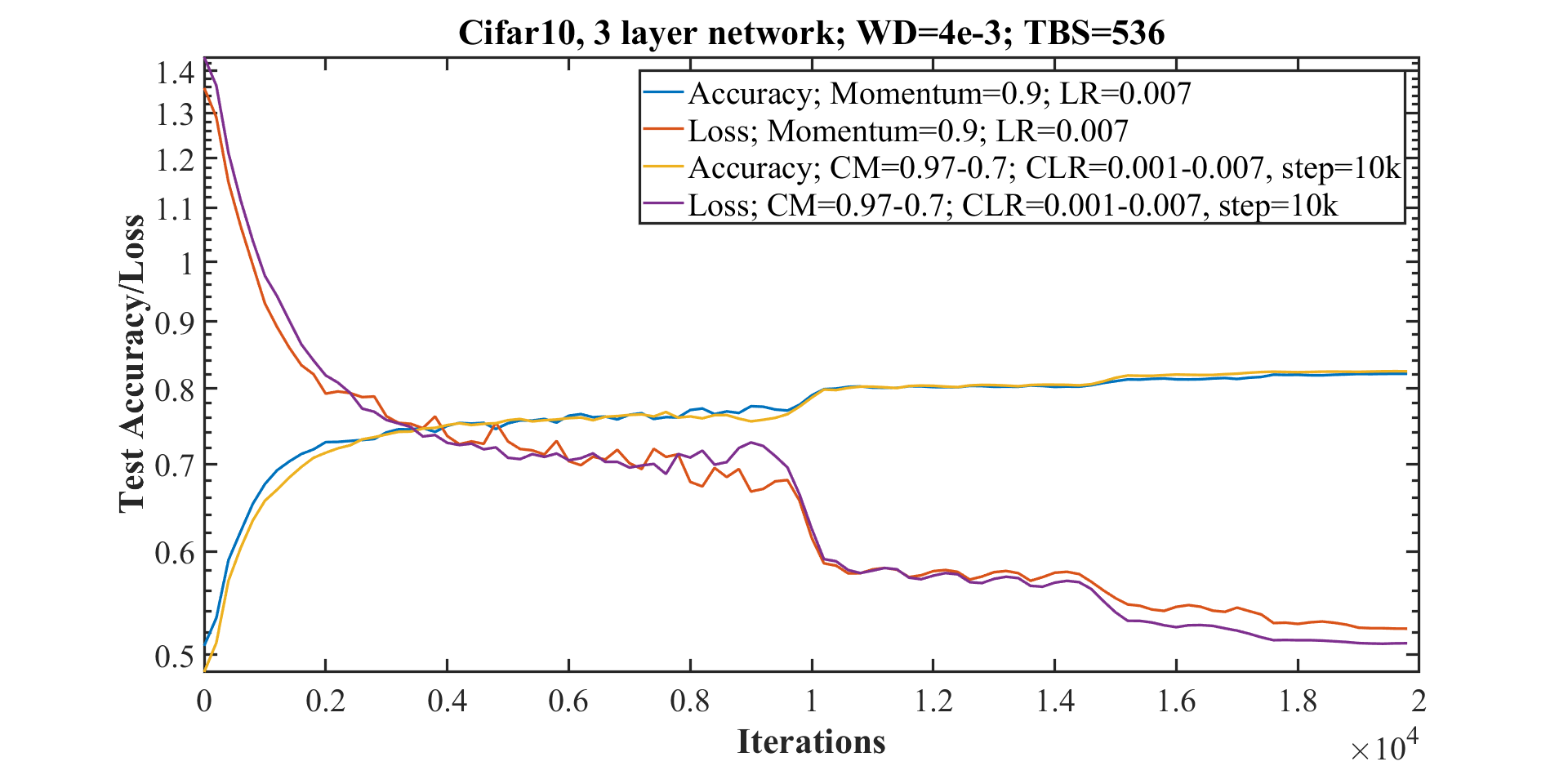}
		\caption{Comparing using a constant momentum to cyclical momentum.}
		\label{fig:3layerCMtestAccLoss2}       
	\end{subfigure}
	\caption{Cyclical momentum tests for the Cifar-10 dataset with a shallow 3 layer network.}
	\label{fig:cyclicalMomentum2}
	\vspace{-10pt}	
\end{figure}

\subsection{Cyclical momentum}
\label{sec:CM}

Momentum and learning rate are closely related.  The optimal learning rate is dependent on the momentum and momentum is dependent on the learning rate.  Since learning rate is regarded as the most important hyper-parameter to tune \citep{bengio2012practical} then momentum is also important.   Like learning rates, it is valuable to set momentum as large as possible without causing instabilities during training.  Figure \ref{fig:3layerMomCLRTestLoss} illustrates the importance of momentum with a 3-layer network on the Cifar-10 dataset.  In this example, a momentum of 0.9 is best and the optimal choice for learning rate clearly depends on the momentum.  Please note that the yellow curve (i.e., momentum = 0.99) displays signs of overfitting (the upward slant after the minimum loss) before diverging (near the learning rate of 0.01) while the blue curve (i.e., momentum = 0.9) does not display overfitting.

Momentum is designed to accelerate network training but its effect on updating the weights is of the same magnitude as the learning rate, as we show here with a brief review of stochastic gradient descent (SGD) with momentum.
In SGD the weights are updated using the negative gradient, given as:
\begin{equation}
\theta_{iter+1} = \theta_{iter} - \epsilon \delta L(F(x,\theta), \theta)
\end{equation}
where $\theta$ represents all the network parameters, $\epsilon$ is the learning rate, and $\delta L(F(x,\theta), \theta)$ is the gradient.  The update rule with momentum is:
\begin{equation}
v_{iter+1} = \alpha v_{iter} - \epsilon \delta L(F(x,\theta), \theta)
\end{equation}
\begin{equation}
\theta_{iter+1} = \theta_{iter} + v
\end{equation}
where $v$ is velocity and $\alpha$ is the momentum coefficient.  From these equation it is apparent that momentum has a similar impact on the weight updates as the learning rate and the velocity is a moving average of the gradient.

While cyclical learning rates and in particular the learning rate range test \citep{smith2017cyclical} are useful methods to find an optimal learning rate, experiments show that a momentum range test is not useful for finding an optimal momentum as can be seen in Figure \ref{fig:3layerCMtestAccLoss}.  In this figure the test loss continues to decrease and accuracy increases as the momentum increases from 0.7 to 1 without an indication of an optimal momentum.  So this begs the question:  \emph{is cyclical momentum useful and if so, when?} 

\textbf{Remark 5.}
The main point of this Section is that \emph{optimal momentum value(s) will improve network training. }  As demonstrated below, the optimal training procedure is a combination of an increasing cyclical learning rate, where an initial small learning rate permits convergence to begin,  and a decreasing cyclical momentum, where the decreasing momentum allows the learning rate to become larger in the early to middle parts of training.  However, if a constant learning rate is used then a large constant momentum (i.e., 0.9-0.99) will act like a pseudo increasing learning rate and will speed up the training.  However, use of too large a value for momentum causes poor training results that are visible early in the training and this can be quickly tested. 

\begin{table}[tb]
	\begin{center}
		\begin{tabular}{| c | c | c | c | c | c | }
			\hline
			Architecture & LR/SS  & momentum/SS & WD & TBS/Epochs & Accuracy (\%) \\ \hline
			3-layer & 0.0005-0.005/11  & 0.95-0.85/11 &  $3 \times 10^{-3} $ & 128/25 & $ 81.3 \pm 0.1 $   \\ \hline
			3-layer & 0.0005-0.005/11  & 0.9-1.0/11  &  $3 \times 10^{-3} $ & 128/25 & $80.2 \pm 0.1 $  \\ \hline
			3-layer & 0.0005-0.005/11  & 0.85 &  $3 \times 10^{-3} $ & 128/25 & $ 79.5 \pm 0.1 $  \\ \hline
			3-layer & 0.0005-0.005/11  & 0.9 &  $3 \times 10^{-3} $ & 128/25 & $ 80.2 \pm 0.4 $ \\ \hline
			3-layer & 0.0005-0.005/11  & 0.95 &  $3 \times 10^{-3} $ & 128/25 & $ 81.2 \pm 0.2 $ \\ \hline
			\hline   
			3-layer & 0.005  & 0.95-0.85/11 &  $3 \times 10^{-3} $ & 128/25 & $ 80.8 \pm 0.3 $   \\ \hline
			3-layer & 0.005  & 0.9-1.0/11 &  $3 \times 10^{-3} $ & 128/25 & $ 80.9 \pm 0.3 $   \\ \hline
			3-layer & 0.005 & 0.85  &  $3 \times 10^{-3} $ & 128/25 & $ 80.2 \pm 0.3 $ \\ \hline
			3-layer & 0.005 & 0.9  &  $3 \times 10^{-3} $ & 128/25 & $ 81.0 \pm 0.2 $ \\ \hline
			3-layer & 0.005 & 0.95 &  $3 \times 10^{-3} $ & 128/25 & $ 81.0 \pm 0.1 $ \\ \hline
			\hline
			resnet-56  & 0.08-0.8/41  & 0.95-0.8/41 &  $ 10^{-4} $ & 512/95 & $ 92.0 \pm 0.2 $   \\ \hline
			resnet-56  & 0.08-0.8/41   & 0.9-1/41 & $ 10^{-4} $ & 512/95 & $ 91.4 \pm 0.1 $   \\ \hline
			resnet-56  & 0.08-0.8/41   & 0.85 & $ 10^{-4} $  & 512/95 & $ 90.8 \pm 0.3 $   \\ \hline
			resnet-56  & 0.08-0.8/41   & 0.9 & $ 10^{-4} $ & 512/95 & $ 91.4 \pm 0.3 $   \\ \hline
			resnet-56  & 0.08-0.8/41   & 0.95 & $ 10^{-4} $ & 512/95 & $ 92.1 \pm 0.1 $   \\ \hline
			\hline
			resnet-56  & 0.1  & 0.95-0.85/41 & $ 10^{-4} $ & 512/95 & $ 89.1 \pm 0.3 $   \\ \hline
			resnet-56  & 0.1  & 0.9-1/41 & $ 10^{-4} $ & 512/95 & $ 88.1 \pm 0.5 $   \\ \hline
			resnet-56  & 0.1  & 0.85 & $ 10^{-4} $ & 512/95 & $ 87.8 \pm 0.3 $   \\ \hline
			resnet-56  & 0.1  & 0.9 & $ 10^{-4} $ & 512/95 & $ 88.1 \pm 0.1 $   \\ \hline
			resnet-56  & 0.1  & 0.95 & $ 10^{-4} $ & 512/95 & $ 88.8 \pm 0.3 $   \\ \hline
		\end{tabular}
		\caption{Cyclical momentum tests; final accuracy and standard deviation for the Cifar-10 dataset with various architectures. For deep architectures, such as resnet-56, combining cyclical learning and cyclical momentum is best, while for shallow architectures optimal constant values work as well as cyclical ones.   SS = stepsize, where two steps in a cycle in epochs, WD = weight decay. }
		\label{tab1:CMcifar}
		\vspace{-20pt}
	\end{center}
	\vspace{-5pt}
\end{table}

Figure \ref{fig:3layerCMtestLoss} shows a LR range test for a shallow, 3-layer architecture on Cifar-10 with the learning rate increasing from 0.002 to 0.02.  The constant momentum case is shown as the blue curve.  The red curve combines the increasing learning rate with a linearly increasing momentum in the range of 0.8 to 1.0.  Although the increasing momentum stabilizes the convergence to a larger learning rate, the minimum test loss is higher than the minimum test loss for the constant momentum case.  On the other hand, decreasing the momentum while the learning rate increases provides three benefits: (1) a lower minimum test loss as shown by the yellow and purple curves, (2) faster initial convergence, as shown by the yellow and purple curves, and (3) greater convergence stability over a larger range of learning rates, as shown by the yellow curve.

This result is reinforced if one compares training the network with a constant momentum (0.9) and learning rate (0.007) versus a cyclical momentum but constant learning rate (step learning rate policy), as illustrated in Figure \ref{fig:3layerCMtestAccLoss2}.  In this run, the average total batch size is 536 (each line is the average of four runs and each run has a slightly different TBS) and the learning rate drops by a factor or 0.316 at iterations 10,000, 15,000, 17,500, and 19,000.  The improvement is clearer with the test loss than with the test accuracy.  Table \ref*{tab1:CMcifar} presents the final accuracies along with the standard deviation.  

Using a decreasing cyclical momentum when the learning rate increases provides an equivalent result to the best constant momentum value but stabilizes the training to allow larger learning rates.  Also note in Figure \ref{fig:3layerCMtestAccLoss2} that the test accuracy for the constant momentum case doesn't plateau, indicating some degree of underfitting while the cyclical case plateaus before iterations 10,000 and 20,000.  The point is that cyclical momentum is useful for starting with a large momentum and decreasing momentum while the learning rate is increasing because it improves the test accuracy and makes the training more robust to large learning rates.  In addition, implementing cyclical momentum is straightforward and example code for implementing cyclical momentum in Caffe is given in the appendix.

A larger value of momentum will speed up the training as will a larger learning rate but can also threaten the stability and cause divergence.  Hence it is useful to choose the momentum wisely.  In addition, a recent paper \citep{liu2018toward} showed that a large momentum helps escape saddle points but can hurt the final convergence, implying that momentum should be reduced at the end of training.  We tested cycling momentum versus decreasing momentum throughout training but found a small improvement with cycling over only decreasing the momentum.

If one is using a cyclical learning rate, a cyclical momentum in the opposite direction makes sense but what is the best momentum when the learning rate is constant? Here cyclical momentum is not better than a good constant value. With either cyclical learning rate or constant learning rate, a good procedure is to test momentum values in the range of 0.9 to 0.99  and choose a value that performs best.  As can be seen in Table \ref{tab1:CMcifar}, a cyclical momentum of 0.95-0.85 provides an equivalent result as to the optimal choice of 0.95, which is better than the accuracy results from using a lower value.  

\begin{figure} [tbh]
	\centering
	\centering
	\begin{subfigure}[b]{0.43\textwidth}
		\includegraphics[width=\textwidth]{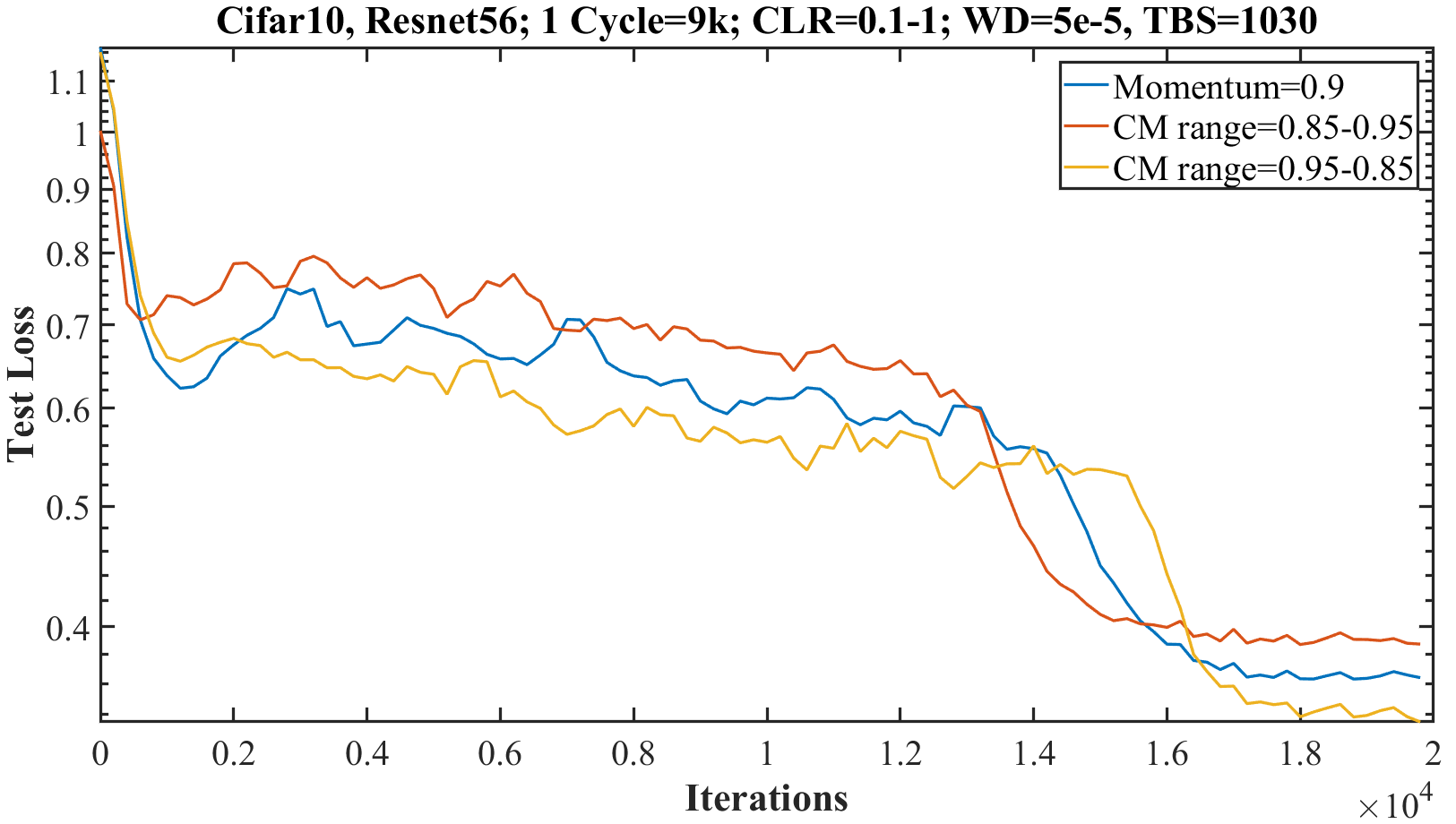}
		\caption{Constant momentum versus two forms for cyclical momentum.}
		\label{fig:resnetCifarSCtestLoss2}       
	\end{subfigure}
	\quad
	\hfill
	~ 
	\begin{subfigure}[b]{0.5\textwidth}
		\includegraphics[width=\textwidth]{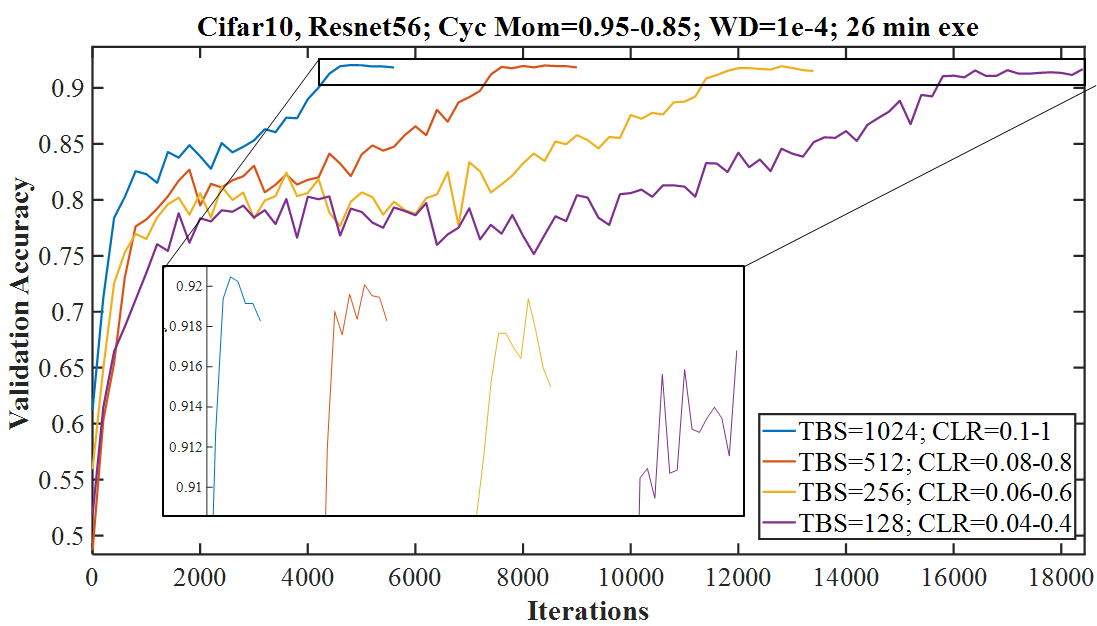}
		\caption{Same as Figure \ref{fig:resnet56CifarTBSAcc} but with cyclical momentum. }
		\label{fig:resnet56CifarTBSAcc2}       
	\end{subfigure}
	\caption{Examples of the cyclical momentum with the Cifar-10 dataset and resnet-56.  The best result is when momentum decreases as learning rate increases. }
	\label{fig:resnetCifarCM}
	\vspace{-5pt}	
\end{figure}


Another reasonable question is \emph{if these lessons carry over to a deeper network, such as resnet-56.}   In this case, Figure \ref{fig:resnetCifarSCtestLoss2} shows a test of cyclical momentum on resnet-56.  This run takes advantage of the fast convergence at large learning rates by cycling up from 0.1 to 1.0 in 9,000 iterations, then down to 0.001 at iteration 18,000.   The same conclusions can be drawn here; that is, decreasing the momentum while increasing the learning rate produces a better performance than a constant momentum or increasing the momentum with the learning rate.  In addition, Figure \ref{fig:resnet56CifarTBSAcc2} is the same experiment as in Figure \ref{fig:resnet56CifarTBSAcc} but here cyclical momentum is used.  Hence, in our experiments, all the general lessons learned from  the shallow network carried over to the deep networks, although the details (i.e., specific values for momentum) varied.  It is good to keep this in mind when initiating a new project or applications.


\begin{figure} [tbh]
	\centering
	\begin{subfigure}[b]{0.5\textwidth}
		\includegraphics[width=\textwidth]{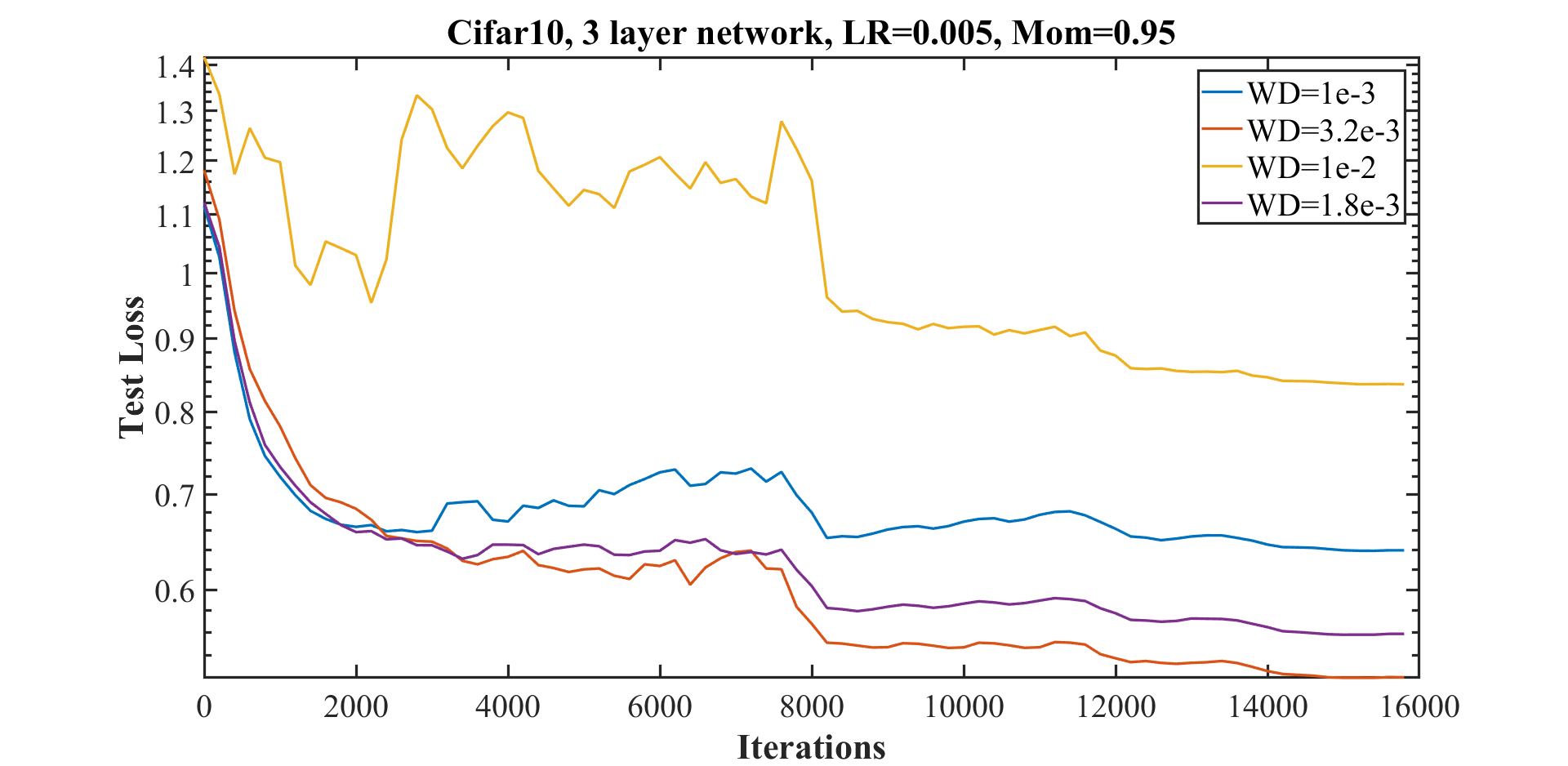}
		\caption{Comparing WD by test loss.}
		\label{fig:3layerWDTestLoss}       
	\end{subfigure}
	\quad
	\hfill
	~ 
	\centering
	\begin{subfigure}[b]{0.43\textwidth}
		\includegraphics[width=\textwidth]{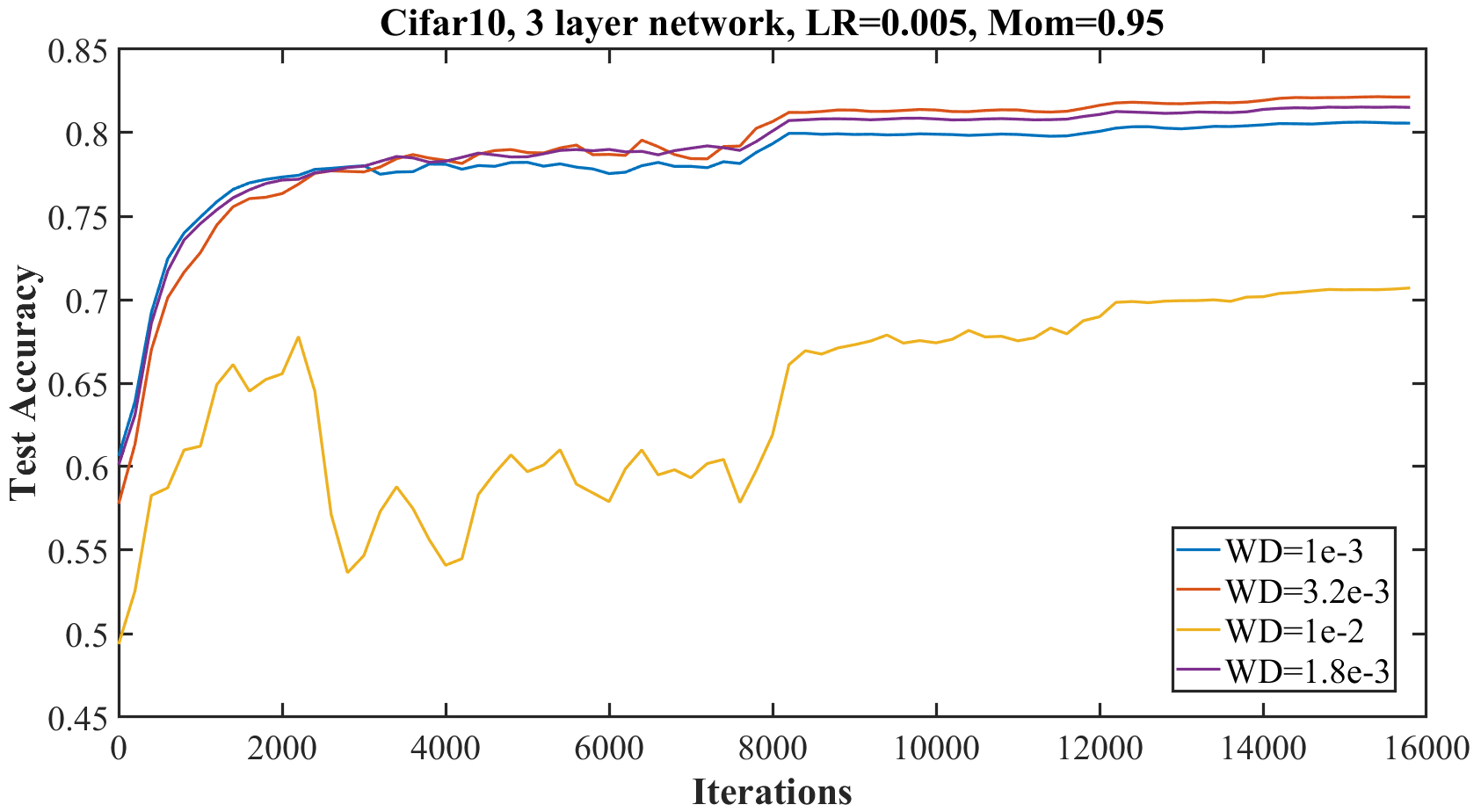}
		\caption{Comparing WD by test accuracy.}
		\label{fig:3layerWDTestAcc}       
	\end{subfigure}
	\caption{Examples of weight decay search using a 3-layer network on the Cifar-10 dataset.  Training used a constant learning rate (0.005) and constant momentum (0.95).  The best value for weight decay is easier to interpret from the loss than from the accuracy.}
\label{fig:3layerWD}
\vspace{-5pt}	
\end{figure}

\subsection{Weight decay}
\label{sec:WDreg}

Weight decay is one form of regularization and it plays an important role in training so its value needs to be set properly.  The important point made above applies; that is, practitioners must balance the various forms of regularization to obtain good performance.  The interested reader can see \cite{kukavcka2017regularization} for a review of regularization methods.  

Our experiments show that weight decay is not like learning rates or momentum and the best value should remain constant through the training (i.e., cyclical weight decay is not useful). This appears to be generally so for regularization but was not tested for all regularization methods (a more complete study of regularization is planned for Part 2 of this report).  Since the network's performance is dependent on a proper weight decay value, a grid search is worthwhile and differences are visible early in the training.  That is, the validation loss early in the training is sufficient for determining a good value.  As shown below, a reasonable procedure is to make combined CLR and CM runs at a few values of the weight decay in order to simultaneously determine the best learning rates, momentum and weight decay. 

If you have no idea of a reasonable value for weight decay, test $ 10^{-3},  10^{-4},  10^{-5},$ and 0. Smaller datasets and architectures seem to require larger values for weight decay while larger datasets and deeper architectures seem to require smaller values.  Our hypothesis is that complex data provides its own regularization and other regularization should be reduced.    

On the other hand, if your experience indicates that a weight decay value of $10^{-4}$ should be about right, these initial runs might be at $3 \times 10^{-5},  10^{-4}, 3 \times 10^{-4}$.  The reasoning behind choosing 3 rather than 5 is that a magnitude is needed for weight decay so this report suggests bisection of the exponent rather than bisecting the value (i.e., between $10^{-4}$ and $10^{-3}$ one bisects as $ 10^{-3.5} = 3.16 \times 10^{-4}$ ). Afterwards, make a follow up run that bisects the exponent of the best two of these or if none seem best, extrapolate towards an improved value.

\textbf{Remark 6.}
Since  \emph{the amount of regularization must be balanced for each dataset and architecture,} the value of weight decay is a key knob to turn for tuning regularization against the regularization from an increasing learning rate.  While other forms of regularization are generally fixed (i.e., dropout ratio, stochastic depth), one can easily change the weight decay value when experimenting with maximum learning rate and stepsize values.

Figure \ref{fig:3layerWDTestLoss} shows the validation loss of a grid search for a 3-layer network on Cifar-10 data, after assuming a learning rate of 0.005 and momentum of 0.95.  Here it would be reasonable to run values of $1 \times 10^{-2}, 3.2 \times  10^{-3}, 10^{-3}$, which are shown in the Figure.  Clearly the yellow curve implies that $1 \times 10^{-2} $ is too large and the blue curve implies that $10^{-3}$ is too small (notice the overfitting).  After running these three, a value of $ 3.2 \times  10^{-3}$ seems right but one can also make a run with a weight decay value of  $10^{-2.75} = 1.8 \times  10^{-3}$, which is the purple curve.  This confirms  that $ 3.2 \times  10^{-3}$ is a good choice.   Figure \ref{fig:3layerWDTestAcc} shows the accuracy results from trainings at all four of these values and it is clear that the validation loss is predictive of the best final accuracy.

\begin{figure} [tbh]
	\centering
	\begin{subfigure}[b]{0.5\textwidth}
		\includegraphics[width=\textwidth]{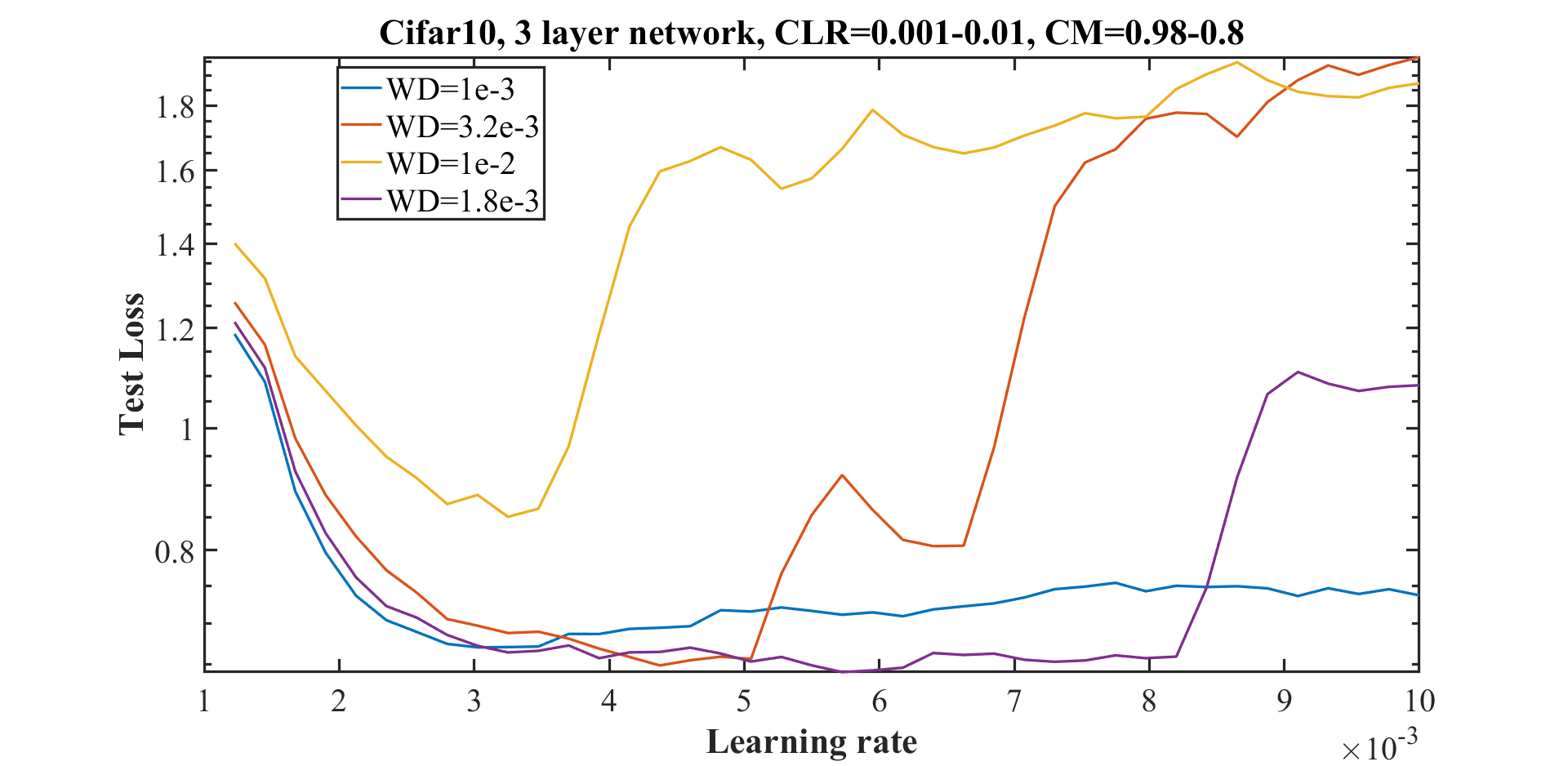}
		\caption{Comparing WD by test loss.}
		\label{fig:3layerWDCLRCMTestLoss}       
	\end{subfigure}
	\quad
	\hfill
	~ 
	\centering
	\begin{subfigure}[b]{0.43\textwidth}
		\includegraphics[width=\textwidth]{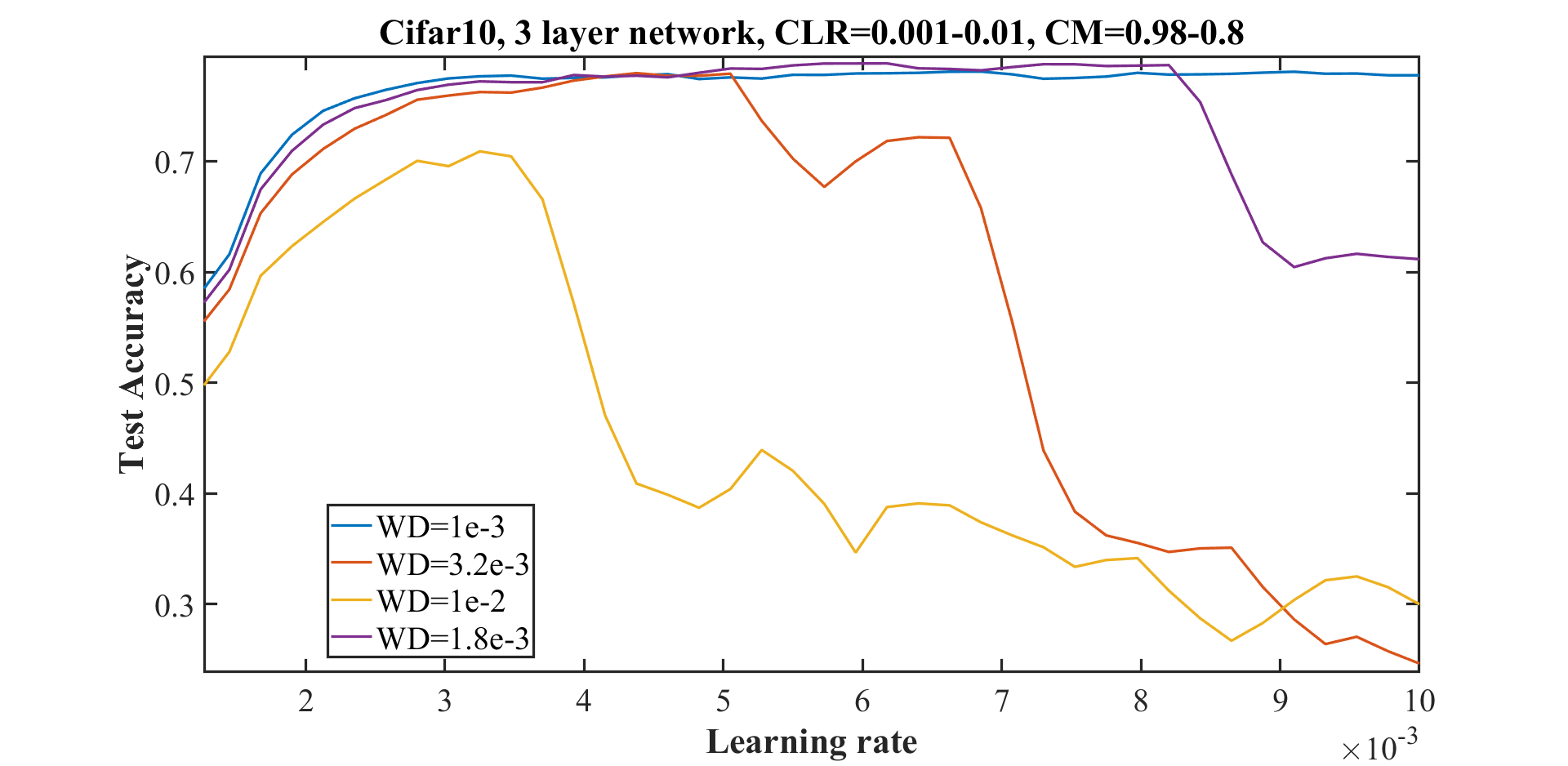}
		\caption{Comparing WD by test accuracy.}
		\label{fig:3layerWDCLRCMTestAcc}       
	\end{subfigure}
	\caption{More examples of weight decay search using a 3-layer network on the Cifar-10 dataset. Training used cyclical learning rates (0.001 - 0.01) and cyclical momentum (0.98 - 0.9). The best value of weight decay is smaller when using CLR because the larger learning rates help with regularization.}
	\label{fig:3layerWD2}
	\vspace{-5pt}	
\end{figure}

A reasonable question is \emph{can the value for the weight decay, learning rate and momentum all be determined simultaneously?} Figure \ref{fig:3layerWDCLRCMTestLoss} shows the runs of a learning rate range test (LR = 0.001 - 0.01) along with a decreasing momentum (= 0.98 - 0.8) at weight decay values of $ 10^{-2}, 3.2 \times  10^{-3}, 10^{-3}$.  As before, a value of $3.2 \times  10^{-3}$ seems best.  However, a test of weight decay at $1.8 \times  10^{-3}$ shows it is better  because it remains stable for larger learning rates and even attains a slightly lower validation loss.  This is confirmed in Figure \ref{fig:3layerWDCLRCMTestAcc} which shows a slightly improved accuracy at learning rates above 0.005.


\begin{figure} [tbh]
	\centering
	\begin{subfigure}[b]{0.5\textwidth}
		\includegraphics[width=\textwidth]{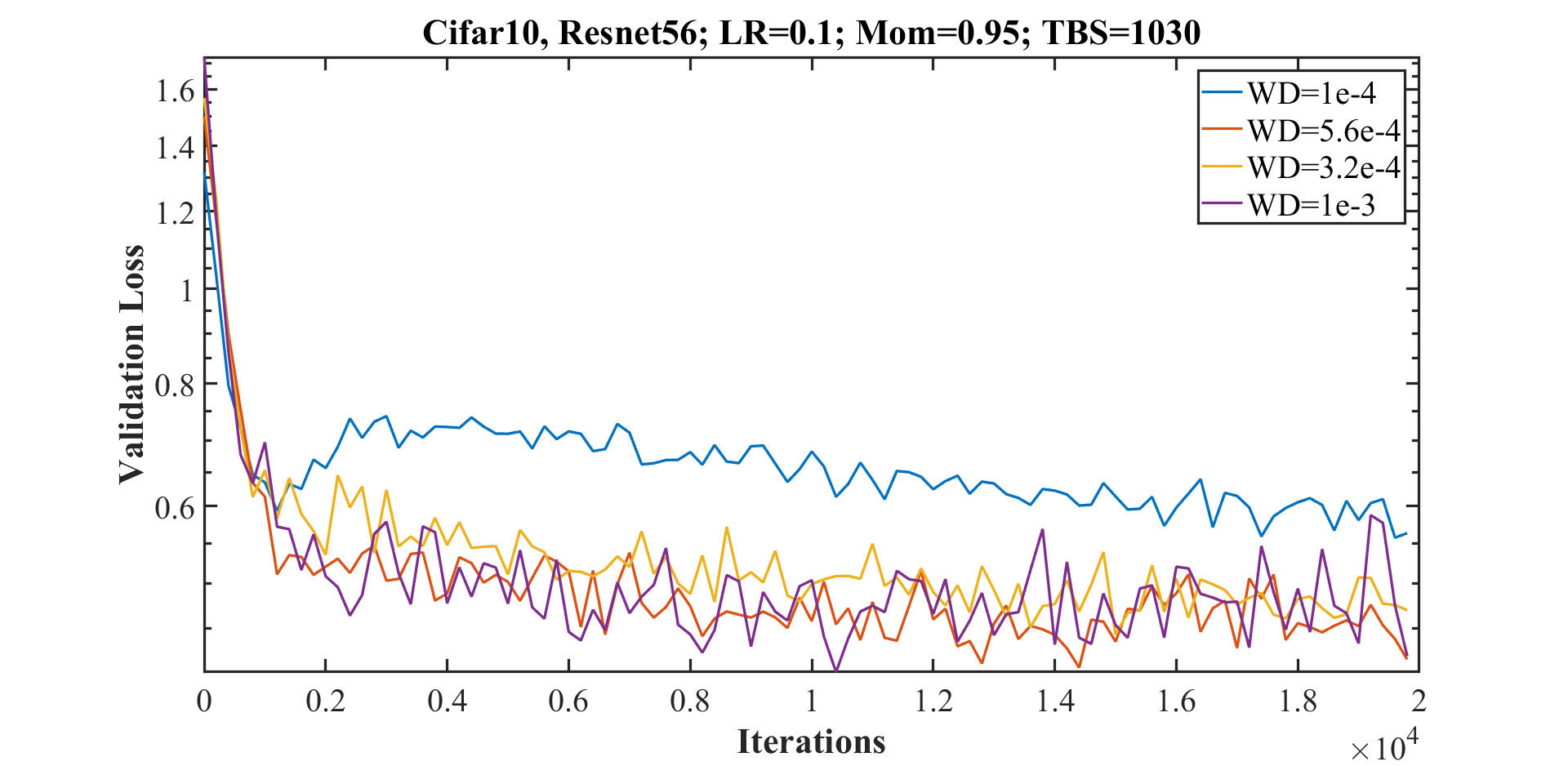}
		\caption{Constant learning rate (=0.1).}
		\label{fig:ResnetCifarWDLRtestLoss}       
	\end{subfigure}
	\quad
	\hfill
	~ 
	\centering
	\begin{subfigure}[b]{0.43\textwidth}
		\includegraphics[width=\textwidth]{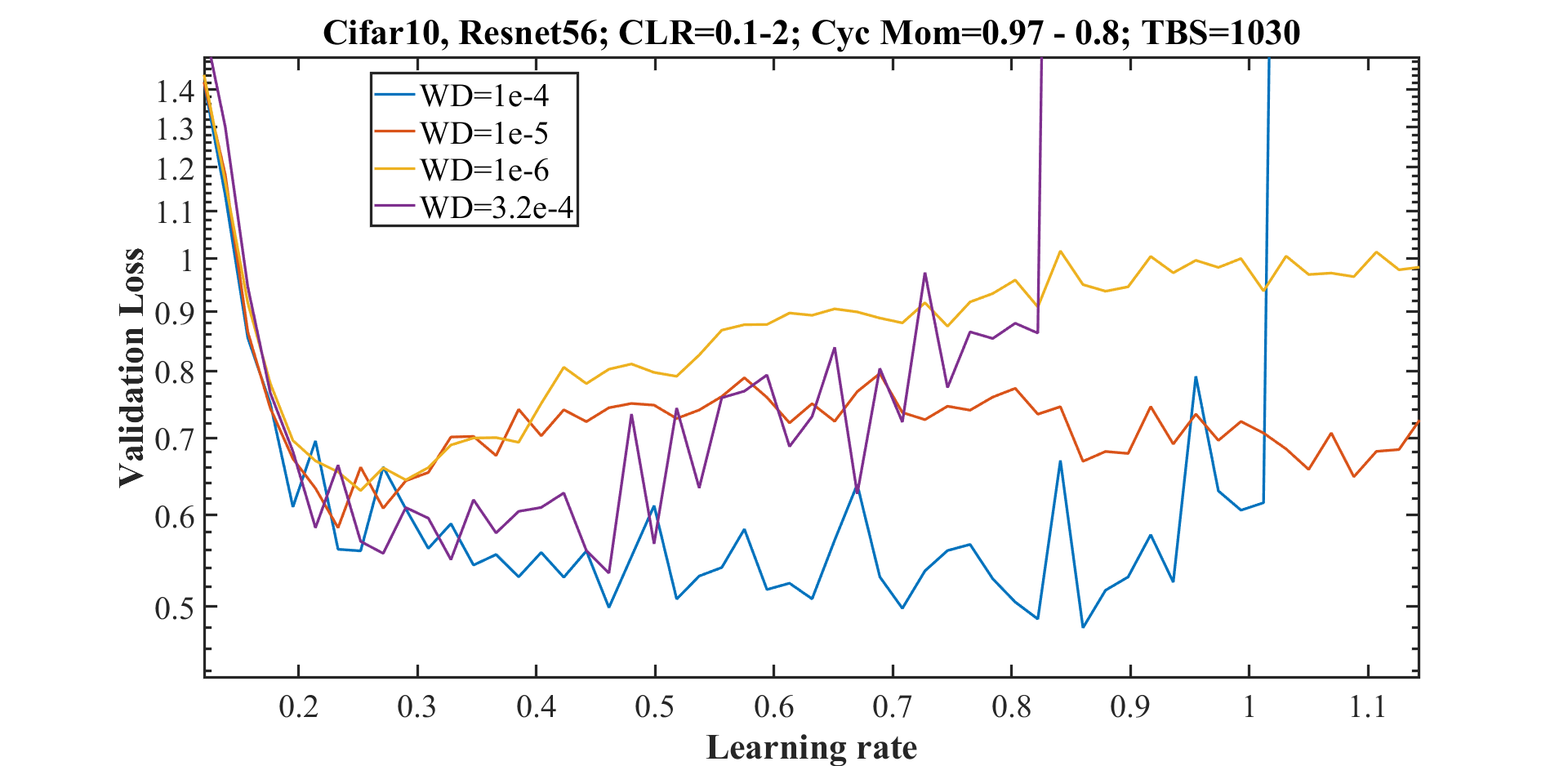}
		\caption{Learning rate range (=0.1-1).}
		\label{fig:ResnetCifarWDCLRtestLoss}       
	\end{subfigure}
	\caption{Grid search for weight decay (WD) on Cifar-10 with resnet-56 and a  constant momentum=0.95 and TBS = 1,030. The optimal weight decay is different if you search with a constant learning rate (left) versus using a learning rate range (right) due to the regularization by large learning rates. }
	\label{fig:ResnetCifarWD}
	\vspace{-5pt}	
\end{figure}

The optimal weight decay is different if you search with a constant learning rate versus using a learning rate range.  This aligns with our intuition because the larger learning rates provide regularization so a smaller weight decay value is optimal.   Figure \ref{fig:ResnetCifarWDLRtestLoss} shows the results of a weight decay search with a constant learning rate of 0.1.  In this case a weight decay of $10^{-4}$ exhibits overfitting and a larger weight decay of $10^{-3}$ is better.  Also shown are the similar results at weight decays of $3.2 \times 10^{-4}$ and $5.6 \times 10^{-4}$ to illustrate that a single significant figure accuracy for weight decay is all that is necessary.   On the other hand, Figure \ref{fig:ResnetCifarWDCLRtestLoss} illustrates the results of a weight decay search using a learning rate range test from 0.1 to 1.0.  This search indicates a smaller weight decay value of $10^{-4}$ is best and larger learning rates in the range of 0.5 to 0.8 should be used. 


\begin{figure} [tbh]
	\centering
	\begin{subfigure}[b]{0.46\textwidth}
		\includegraphics[width=\textwidth]{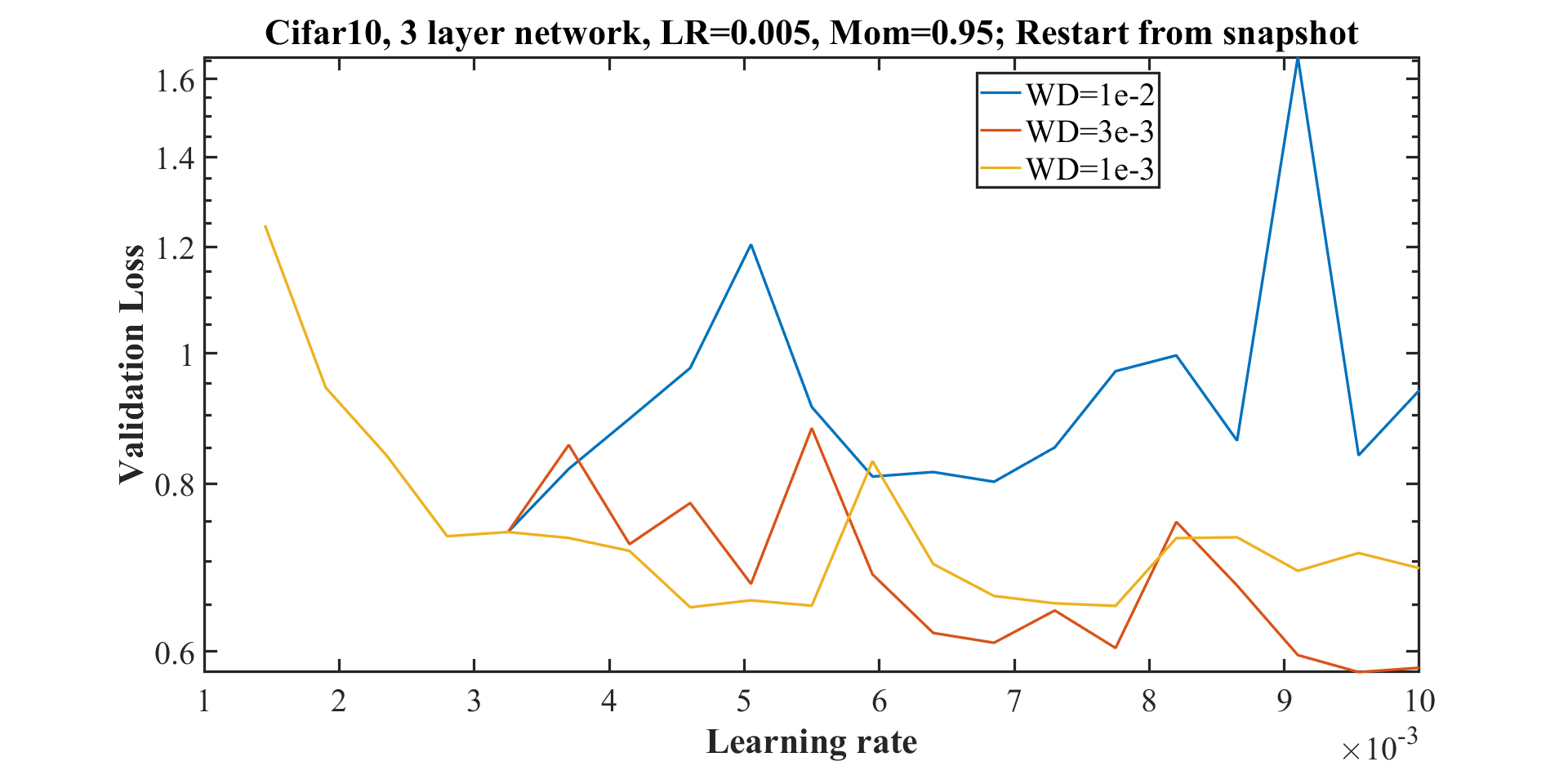}
		\caption{3-layer network.}
		\label{fig:3layerWDsnapshotTestLoss}       
	\end{subfigure}
	\quad
	\hfill
	~ 
	\centering
	\begin{subfigure}[b]{0.46\textwidth}
		\includegraphics[width=\textwidth]{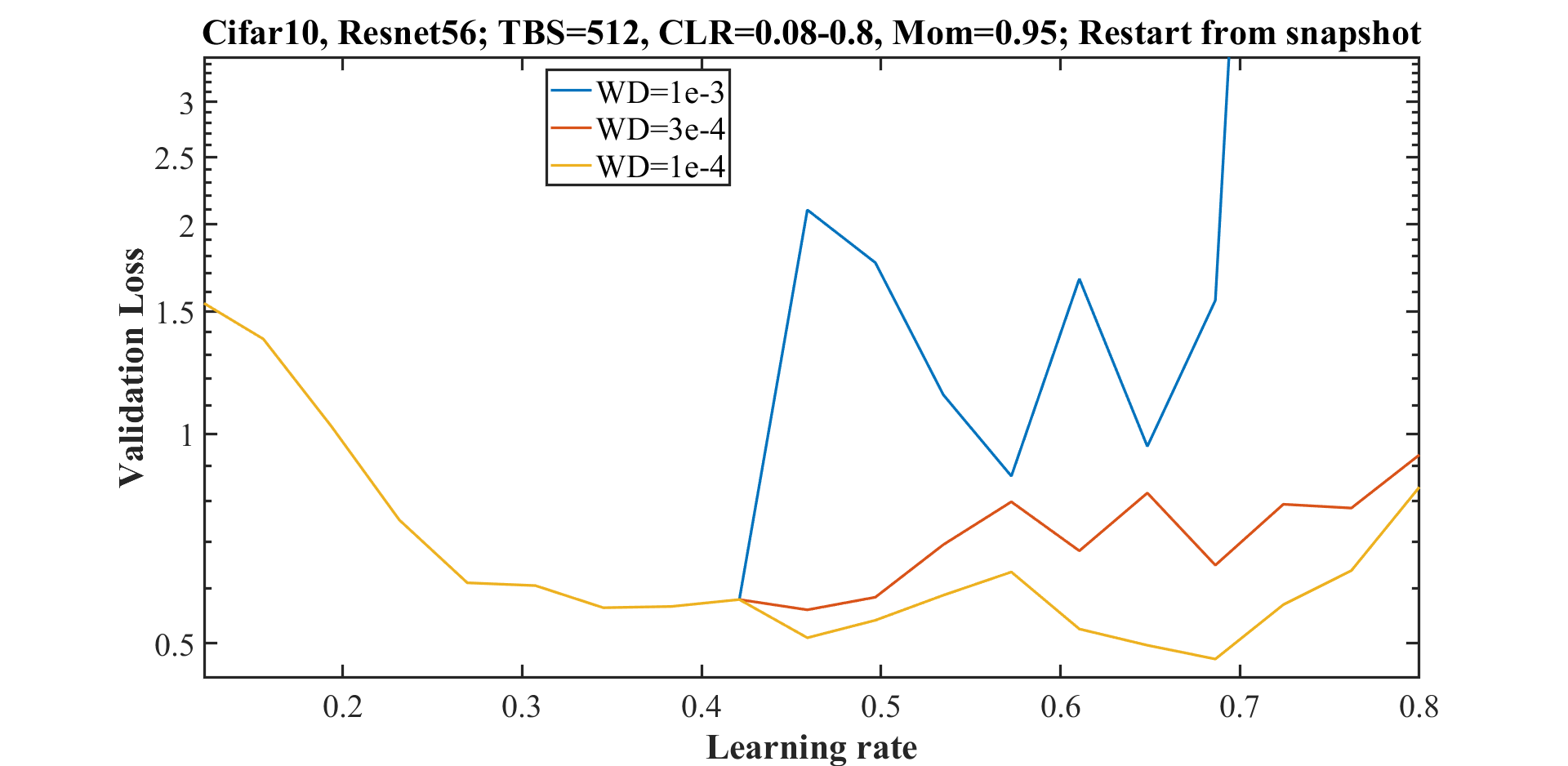}
		\caption{Resnet-56}
		\label{fig:resnet56WDsnapshotTestLoss}       
	\end{subfigure}
	\caption{Grid search for the optimal WD restarting from a snapshot.  These results on Cifar-10 indicate that a grid search from a snapshot will point to an optimal weight decay and can save time. }
	\label{fig:WDfromSnapshot}
	\vspace{-5pt}	
\end{figure}

Another option as a grid search for weight decay is to make a single run at a middle value for weight decay and save a snapshot after the loss plateaus.  Use this snapshot to restart runs, each with a different value of WD.  This can save  time in searching for the best weight decay.  Figure \ref{fig:3layerWDsnapshotTestLoss} shows an example on Cifar-10 with a 3-layer network (this is for illustration only as this architecture runs very quickly).  Here the initial run was with a sub-optimal value of weight decay of $10^{-3}$.  From the restart point, three continuation runs were made with weight decay values of $10^{-3}$, $3 \times 10^{-3}$, and $10^{-2}$. This Figure shows that $3 \times 10^{-3}$ is best.

Figure \ref{fig:resnet56WDsnapshotTestLoss} illustrates a weight decay grid search from a snapshot for resnet-56 while performing a LR range test.  This Figure shows the first half of the range test with a value of weight decay of $10^{-4}$.  Then three continuations are run with weight decay values of $10^{-3}$, $3 \times 10^{-4}$, and $10^{-4}$.  It is clear that a weight decay value of $10^{-4}$ is best and information about the learning rate range is simultaneously available.

\section{Experiments with other architectures and datasets}
\label{sec:other}

All of the above can be condensed into a short recipe for finding a good set of hyper-parameters with a given dataset and architecture.
\begin{enumerate}
	\item Learning rate (LR): Perform a learning rate range test to a ``large'' learning rate.  The max LR depends on the architecture (for the shallow 3-layer architecture, large is 0.01 while for resnet, large is 3.0), you might try more than one maximum.  Using the 1cycle LR policy with a maximum learning rate determined from an LR range test, a minimum learning rate as a tenth of the maximum appears to work well but other factors are relevant, such as the rate of learning rate increase (too fast and increase will cause instabilities).
	\item Total batch size (TBS): A large batch size works well but the magnitude is typically constrained by the GPU memory.  If your server has multiple GPUs, the total batch size is the batch size on a GPU multiplied by the number of GPUs.  If the architecture is small or your hardware permits very large batch sizes, then you might compare performance of different batch sizes.  In addition, recall that small batch sizes add regularization while large batch sizes add less, so utilize this while balancing the proper amount of regularization.  It is often better to use a larger batch size so a larger learning rate can be used.
	\item Momentum: Short runs with momentum values of 0.99, 0.97, 0.95, and 0.9 will quickly show the best value for momentum.  If using the 1cycle learning rate schedule, it is better to use a cyclical momentum (CM) that starts at this maximum momentum value and decreases with increasing learning rate to a value of 0.8 or 0.85 (performance is almost independent of the minimum momentum value).  Using cyclical momentum along with the LR range test stabilizes the convergence when using large learning rate values more than a constant momentum does.
	\item Weight decay (WD): This requires a grid search to determine the proper magnitude but usually does not require more than one significant figure accuracy.  Use your knowledge of the dataset and architecture to decide which values to test.  For example, a more complex dataset requires less regularization so test smaller weight decay values, such as $10^{-4}, 10^{-5}, 10^{-6}, 0$.  A shallow architecture requires more regularization  so test larger weight decay values, such as $10^{-2}, 10^{-3}, 10^{-4}$.
\end{enumerate}  

This list summarizes the techniques described in this report for hyper-parameter optimization. Hyper-parameter optimization can be reasonably quick if one searches for clues in the test loss early in the training.  For illustration, the Sections below describe the use of the above checklist for several other architectures and datasets.  Files to help replicate the results reported here are available at https://github.com/lnsmith54/hyperParam1.

\begin{figure} [tbh]
	\centering
	\begin{subfigure}[b]{0.46\textwidth}
		\includegraphics[width=\textwidth]{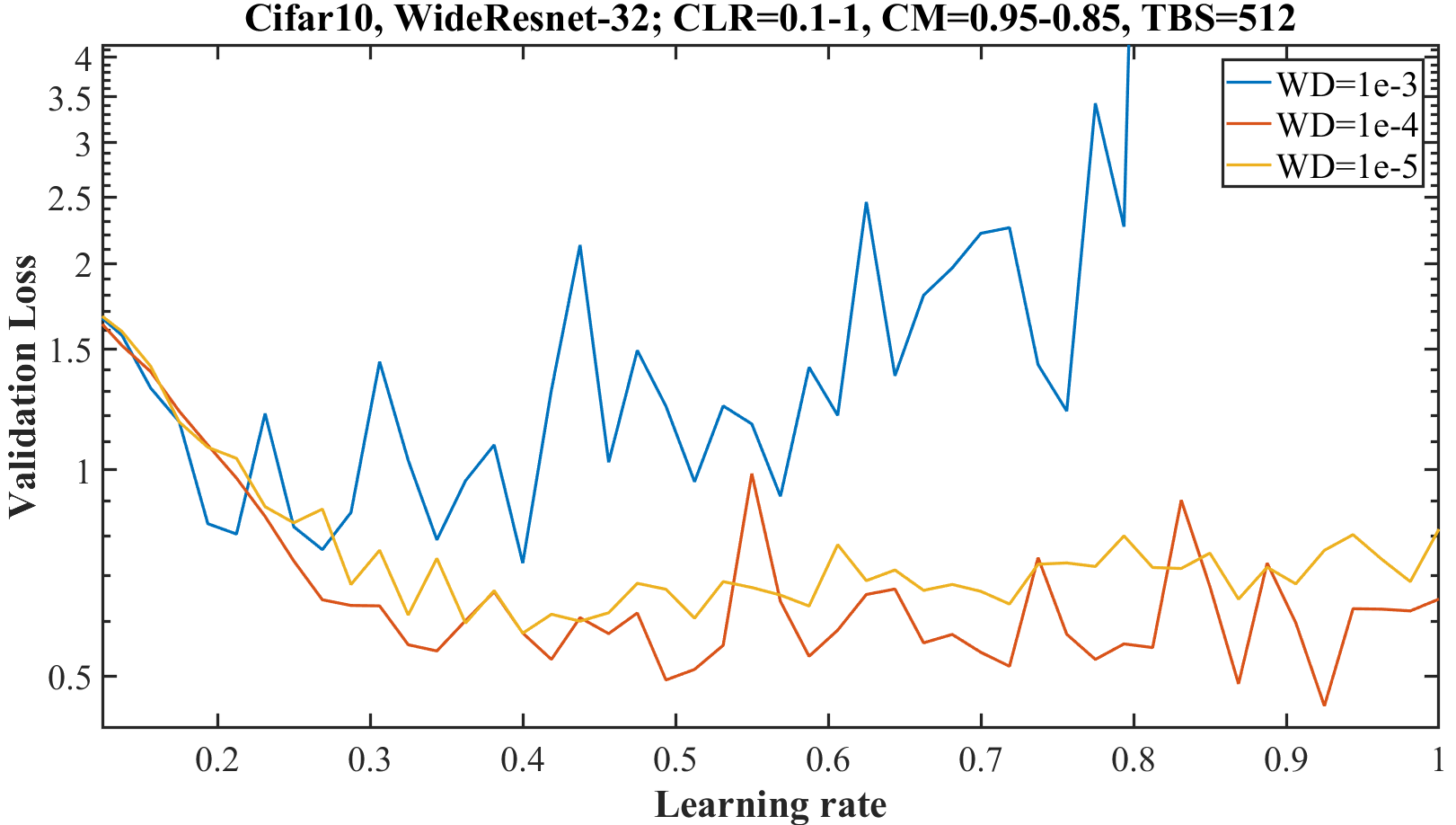}
		\caption{Comparison of WD with a 32 layer wide resnet.}
		\label{fig:wide32cifarWDTestLoss}       
	\end{subfigure}
	\hfill
	~ 
	\centering
	\begin{subfigure}[b]{0.46\textwidth}
		\includegraphics[width=\textwidth]{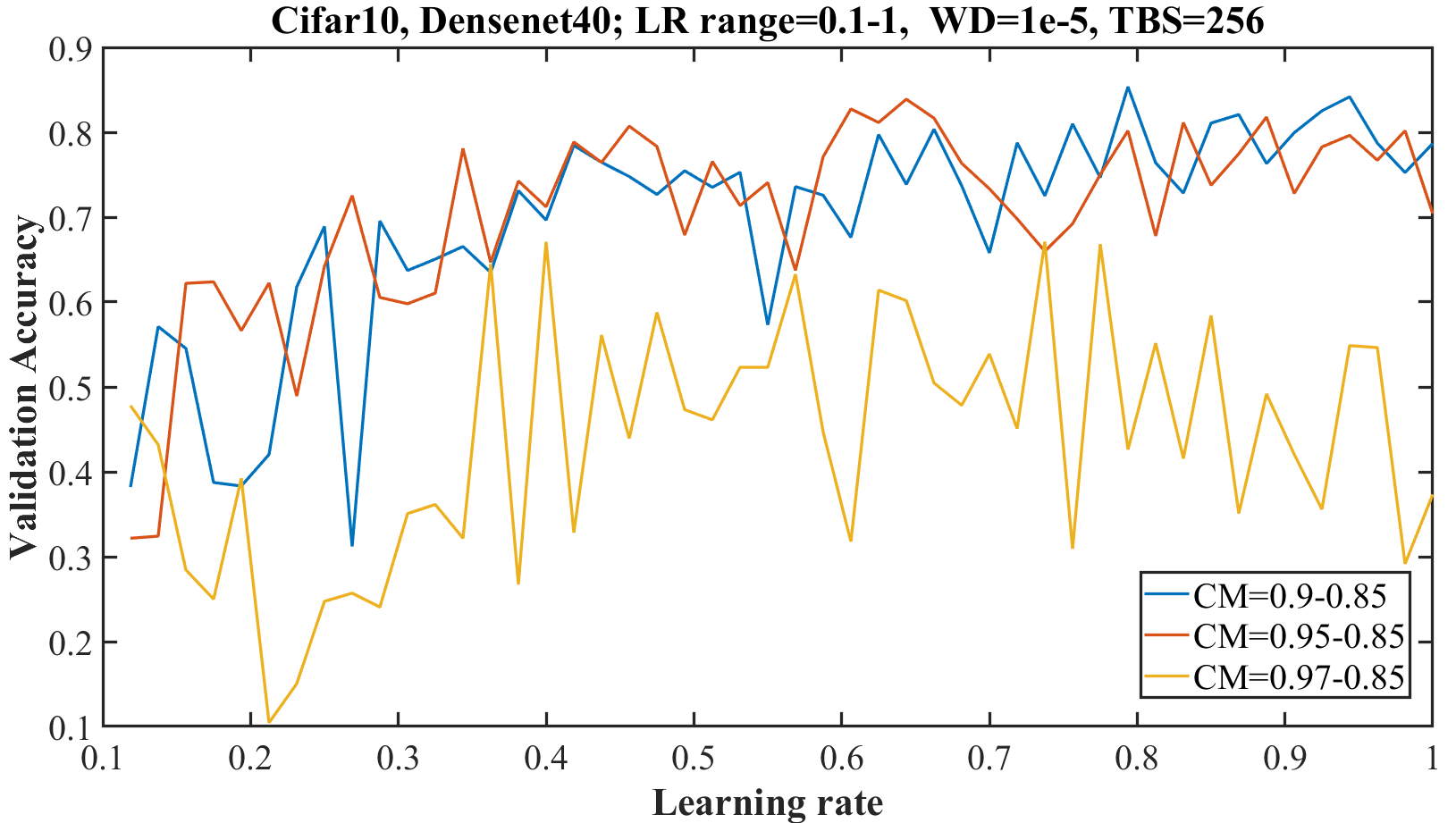}
		\caption{Comparison of momentum with a 40 layer densenet. }
		\label{fig:dense40cifarCM}       
	\end{subfigure}
	\quad
	\hfill
	\centering
	\begin{subfigure}[b]{0.46\textwidth}
		\includegraphics[width=\textwidth]{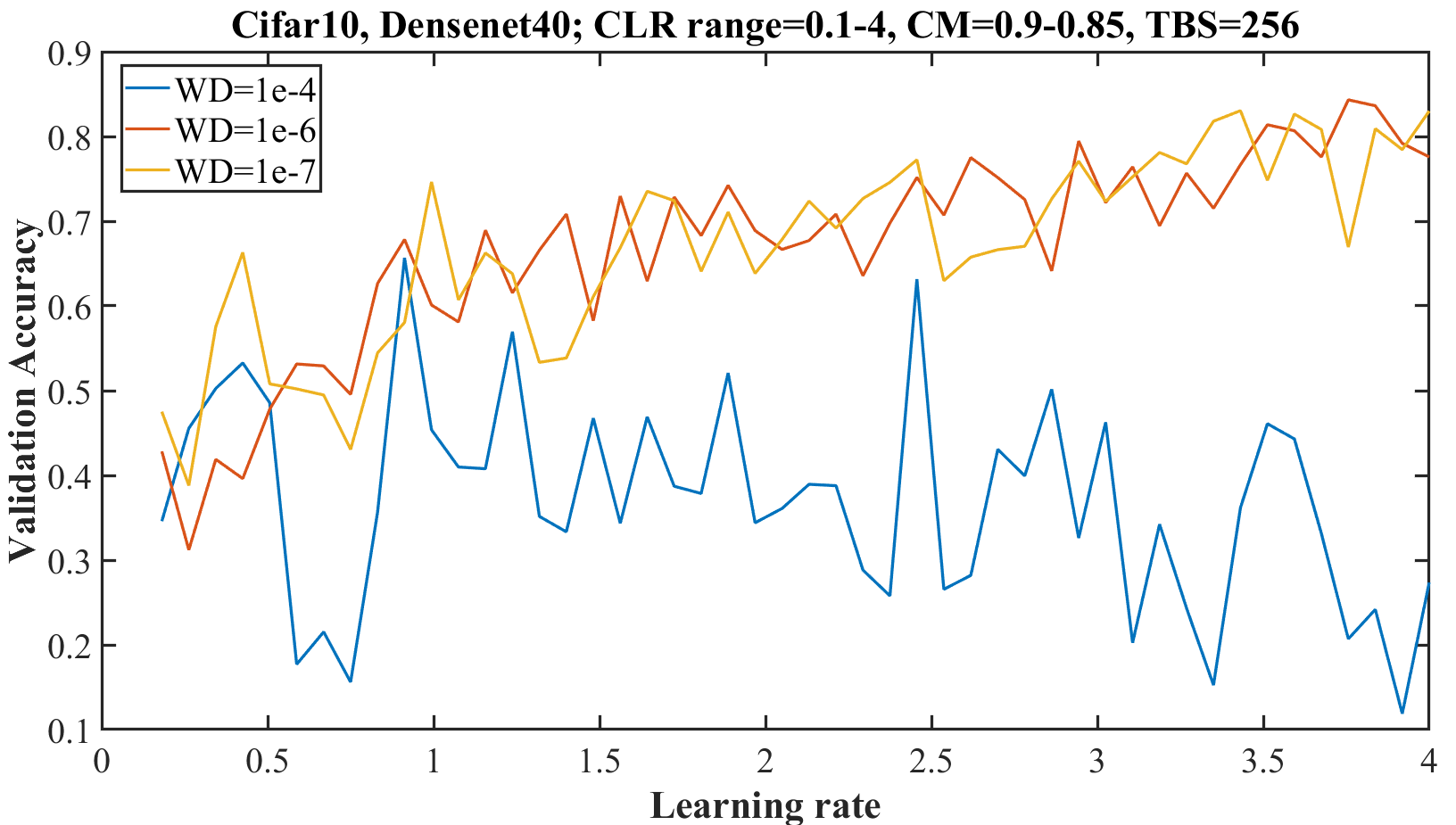}
		\caption{Comparison of WD with 40 layer densesnet.}
		\label{fig:dense40cifarWD}       
	\end{subfigure}
	\hfill
	\centering
	\begin{subfigure}[b]{0.46\textwidth}
		\includegraphics[width=\textwidth]{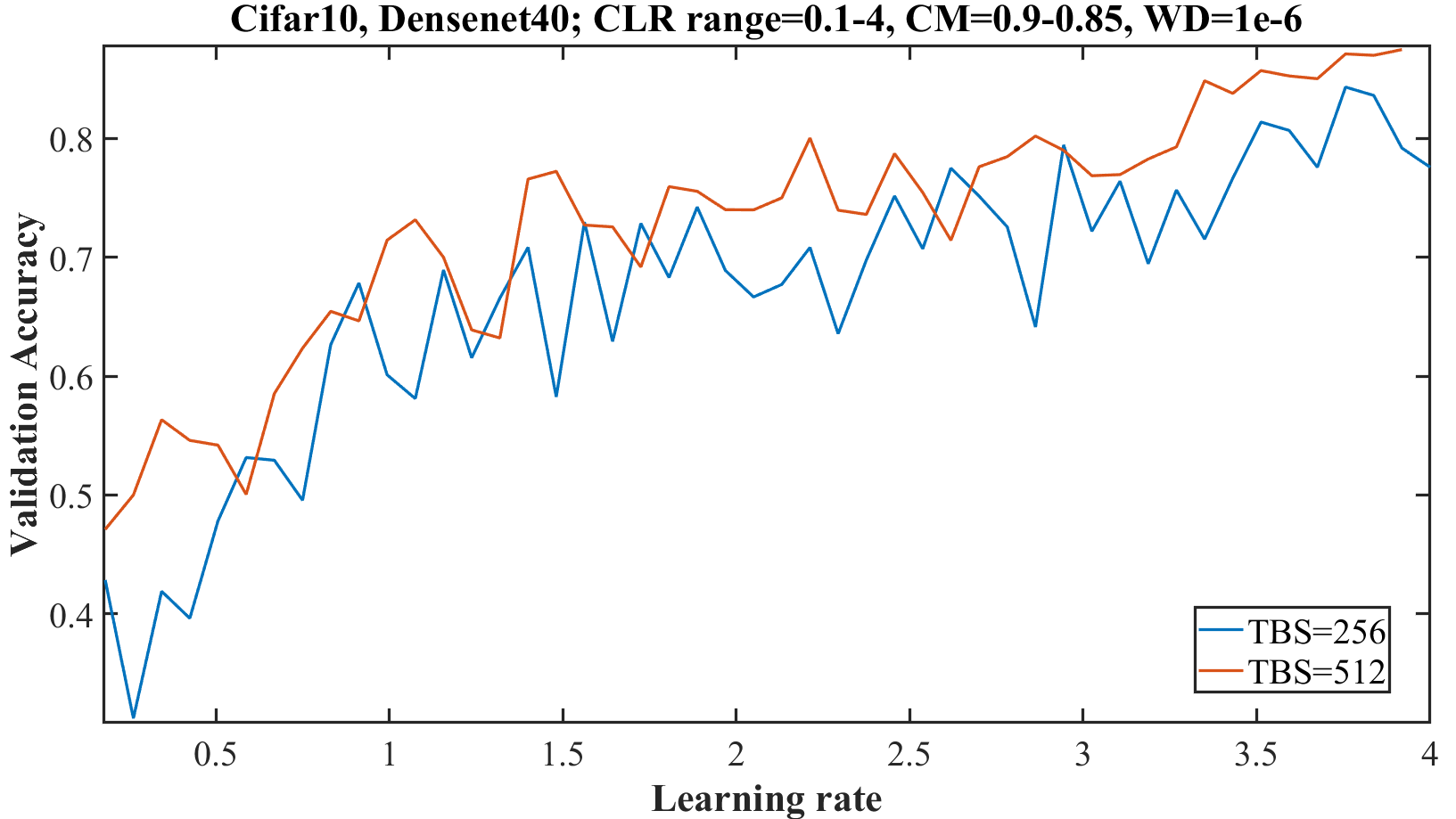}
		\caption{Comparison of TBS with 40 layer densenet. }
		\label{fig:dense40cifarTBS}       
	\end{subfigure}
	~ 
	\caption{Illustration of hyper-parameter search for wide resnet and densenet on Cifar-10.  Training follows the learning rate range test  of (LR=0.1 -- 1) for widenet and (LR=0.1 -- 4) for densenet, and cycling momentum (=0.95 -- 0.85).  For the densenet architecture, the test accuracy is easier to interpret for the best weight decay (WD)  than the test loss.}
	\label{fig:WideDensenet}
	\vspace{-5pt}	
\end{figure}

\subsection{Wide resnets on Cifar-10}
\label{sec:wide}

The wide resnet was created from a resnet with 32 layers by increasing the number of channels by a factor of 4 instead of the factor of 2 used by resnet.  

The first steps for testing the hyper-parameters with the wide resnet architecture is to run a LR range test with decreasing momentum at a few values of weight decay.  Since wide resnets are similar to resnet, we assumed a TBS = 512, a learning rate range from 0.1 to 1.0, and a momentum range from 0.95 to 0.85.  Figure \ref{fig:wide32cifarWDTestLoss} illustrates a grid search on weight decay with three runs, each with different weight decay values of $10^{-3}, 10^{-4}$, and $10^{-5}$.  This Figure shows that $10^{-3}$ yields poor performance and $10^{-4}$ is a little better than $10^{-5}$.  A fourth test with weight decay set to $3 \times 10^{-5}$ (not shown) is similar to both $10^{-4}$ and $10^{-5}$, indicating any value in this range is good.

Table \ref{tab:otherExamples} provides the final result of training with the discovered hyper-parameters, using the 1cycle learning rate schedule with learning rate bounds from 0.1 to 1.0.  In 100 epochs, the wide32 network converges and provides a test accuracy of $91.9\% \pm 0.2$ while the standard training method achieves an accuracy of only $ 90.3 \pm 1.0 $ in 800 epochs.  This demonstrates super-convergence for wide resnets.

\begin{table}[tb]
	\begin{center}
		\begin{tabular}{| c | c | c | c | c | c | c | }
			\hline
			Dataset & Architecture & CLR/SS/PL  & CM/SS & WD & Epochs & Accuracy (\%) \\ \hline
			Cifar-10 & wide resnet & 0.1/Step  & 0.9  & $10^{-4}$  &  100 & $ 86.7 \pm 0.6 $   \\ \hline
			Cifar-10 & wide resnet & 0.1/Step  & 0.9  & $10^{-4}$  &  200 & $ 88.7 \pm 0.6 $   \\ \hline
			Cifar-10 & wide resnet & 0.1/Step  & 0.9  & $10^{-4}$  &  400 & $ 89.8 \pm 0.4 $   \\ \hline
			Cifar-10 & wide resnet & 0.1/Step  & 0.9  & $10^{-4}$  &  800 & $ 90.3 \pm 1.0 $   \\ \hline
			Cifar-10 & wide resnet & 0.1-0.5/12  & 0.95-0.85/12 & $10^{-4}$ &  25 & $ 87.3 \pm 0.8 $   \\ \hline
			Cifar-10 & wide resnet & 0.1-1/23  & 0.95-0.85/23 & $10^{-4}$  &   50 & $ 91.3 \pm 0.1 $   \\ \hline
			Cifar-10 & wide resnet & 0.1-1/45  & 0.95-0.85/45 & $10^{-4}$  &  100 & $ 91.9 \pm 0.2 $   \\ \hline
			\hline   
			Cifar-10 & densenet & 0.1/Step  & 0.9 & $10^{-4}$  &  100 & $ 91.3 \pm 0.2 $   \\ \hline
			Cifar-10 & densenet & 0.1/Step  & 0.9 & $10^{-4}$  &  200 & $ 92.1 \pm 0.2 $   \\ \hline
			Cifar-10 & densenet & 0.1/Step  & 0.9 & $10^{-4}$  &  400 & $ 92.7 \pm 0.2 $   \\ \hline
			Cifar-10 & densenet & 0.1-4/22  & 0.9-0.85/22 & $10^{-6}$  &  50 & $ 91.7 \pm 0.3 $   \\ \hline
			Cifar-10 & densenet & 0.1-4/34  & 0.9-0.85/34 & $10^{-6}$  &  75 & $ 92.1 \pm 0.2 $   \\ \hline
			Cifar-10 & densenet & 0.1-4/45  & 0.9-0.85/45 & $10^{-6}$  &  100 & $ 92.2 \pm 0.2 $   \\ \hline
			Cifar-10 & densenet & 0.1-4/70  & 0.9-0.85/70 & $10^{-6}$  &  150 & $ 92.8 \pm 0.1 $   \\ \hline
			\hline   
			MNIST  & LeNet & 0.01/inv & 0.9  & $5 \times 10^{-4}$  &  85 & $ 99.03 \pm 0.04 $   \\ \hline
			MNIST  & LeNet & 0.01/step  & 0.9  & $5 \times 10^{-4}$  &  85 & $ 99.00 \pm 0.04 $   \\ \hline
			MNIST  & LeNet & 0.01-0.1/5  & 0.95-0.8/5 & $5 \times 10^{-4}$  &  12 & $ 99.25  \pm 0.03 $   \\ \hline
			MNIST  & LeNet & 0.01-0.1/12  & 0.95-0.8/12 & $5 \times 10^{-4}$  &  25 & $ 99.28  \pm 0.06 $   \\ \hline
			MNIST  & LeNet & 0.01-0.1/23  & 0.95-0.8/23 & $5 \times 10^{-4}$  &  50 & $ 99.27  \pm 0.07 $   \\ \hline
			MNIST  & LeNet & 0.02-0.2/40  & 0.95-0.8/40 & $5 \times 10^{-4}$  &  85 & $ 99.35  \pm 0.03 $   \\ \hline
			\hline   
			Cifar-100 & resnet-56 & 0.005/step & 0.9  & $10^{-4}$  &  100 & $ 60.8 \pm 0.4 $   \\ \hline
			Cifar-100 & resnet-56 & 0.005/step & 0.9  & $10^{-4}$  &  200 & $ 61.6 \pm 0.9 $   \\ \hline
			Cifar-100 & resnet-56 & 0.005/step & 0.9  & $10^{-4}$  &  400 & $ 61.0 \pm 0.2 $   \\ \hline
			Cifar-100 & resnet-56 & 0.1-0.5/12 & 0.95-0.85/12 & $10^{-4}$  &  25 & $ 65.4 \pm 0.2 $   \\ \hline
			Cifar-100 & resnet-56 & 0.1-0.5/23 & 0.95-0.85/23 & $10^{-4}$  &  50 & $ 66.4 \pm 0.6 $   \\ \hline
			Cifar-100 & resnet-56 & 0.09-0.9/45 & 0.95-0.85/45 & $10^{-4}$  &  100 & $ 69.0 \pm 0.4 $   \\ \hline
		\end{tabular}
		\vspace{10pt}
		\caption{Final accuracy and standard deviation for various datasets and architectures.  The total batch size (TBS) for all of the reported runs was 512.  PL = learning rate policy, SS = stepsize in epochs, where two steps are in a cycle, WD = weight decay, CM = cyclical momentum. Either SS or PL is provide in the Table and SS implies the cycle learning rate policy. }
		\label{tab:otherExamples}
	\end{center}
	\vspace{-20pt}
\end{table}

\subsection{Densenets on Cifar-10}
\label{sec:dense}

A 40 layer densenet architecture was create from the code at \url{https://github.com/liuzhuang13/DenseNetCaffe}.   For comparison, values of the hyper-parameters provided at this website are; learning rate = 0.1, momentum = 0.9, and weight decay = 0.0001. 

The same procedure described above can be performed for a 40 layer densenet architecture but finding the hyper-parameters for densenet is more challenging  than with wide resnets.  The first steps for testing the hyper-parameters with the densenet architecture is to run a LR range test with a few maximum values for momentum.  Momentum set to 0.99 diverges but values of 0.97, 0.95, and 0.9 are shown in Figure \ref{fig:dense40cifarCM}.     While it is typically easier to interpret the test loss for finding the best hyper-parameter values, for the densenet architecture the test accuracy is  visibly easier to interpret  than the test loss and is displayed in this Figure.   For the densenet architecture, smaller values of momentum perform better so a range from 0.9 decreasing to 0.85 was used in subsequent tests.  Next the learning rate range was tested.  As shown in Figure \ref{fig:dense40cifarWD}, the densenet architecture is stable even with a learning rate range from 0.1 to 4.0 (we didn't test larger than 4).  

Using a TBS = 256 (an initial guess), a combined  learning rate and a momentum range (from 0.95 to 0.85), three run with different weight decay values of $10^{-3}, 10^{-4}$, and $10^{-5}$ again shows that $10^{-3}$ yields poor performance and $10^{-5}$ is best.  Hence smaller weight decay values were tested and the results are shown in Figure \ref{fig:dense40cifarWD}.  It is clear in this Figure that weight decay of $10^{-4}$ produces poorer accuracy than smaller weight decay values.  In addition, the stochasticity of the curve implies that the architecture complexity is adding regularization so reducing the weight decay makes intuitive sense.  Figure \ref{fig:dense40cifarWD} implies that a weight decay value $10^{-6}$ seems about right.

The next hyper-parameter test shown in Figure \ref{fig:dense40cifarTBS} is for the total batch size (TBS).  This plot compares TBS = 256 to 512 (a TBS of 1024 didn't fit in our server's GPU memory).  This Figure indicates that TBS = 512 performs better than 256.  The larger batch size also reduces the regularization, which is reflected in the slightly less noisy curve.  

Table \ref{tab:otherExamples} provides the final accuracy results of training with the discovered hyper-parameters, using the 1cycle learning rate schedule with learning rate bounds from 0.1 to 4.0 and cyclical momentum in the range of 0.9 to 0.85.  This Table also shows the effects of longer training lengths, where the final accuracy improves from 91.7\% for a quick 50 epoch (4,882 iterations) training to 92.8\% with a longer 150 epoch (14,648 iterations) training.  The step learning rate policy attains an equivalent accuracy of 92.7\% but requires 400 epochs to do so.

\begin{figure} [tbh]
	\centering
	\begin{subfigure}[b]{0.47\textwidth}
		\includegraphics[width=\textwidth]{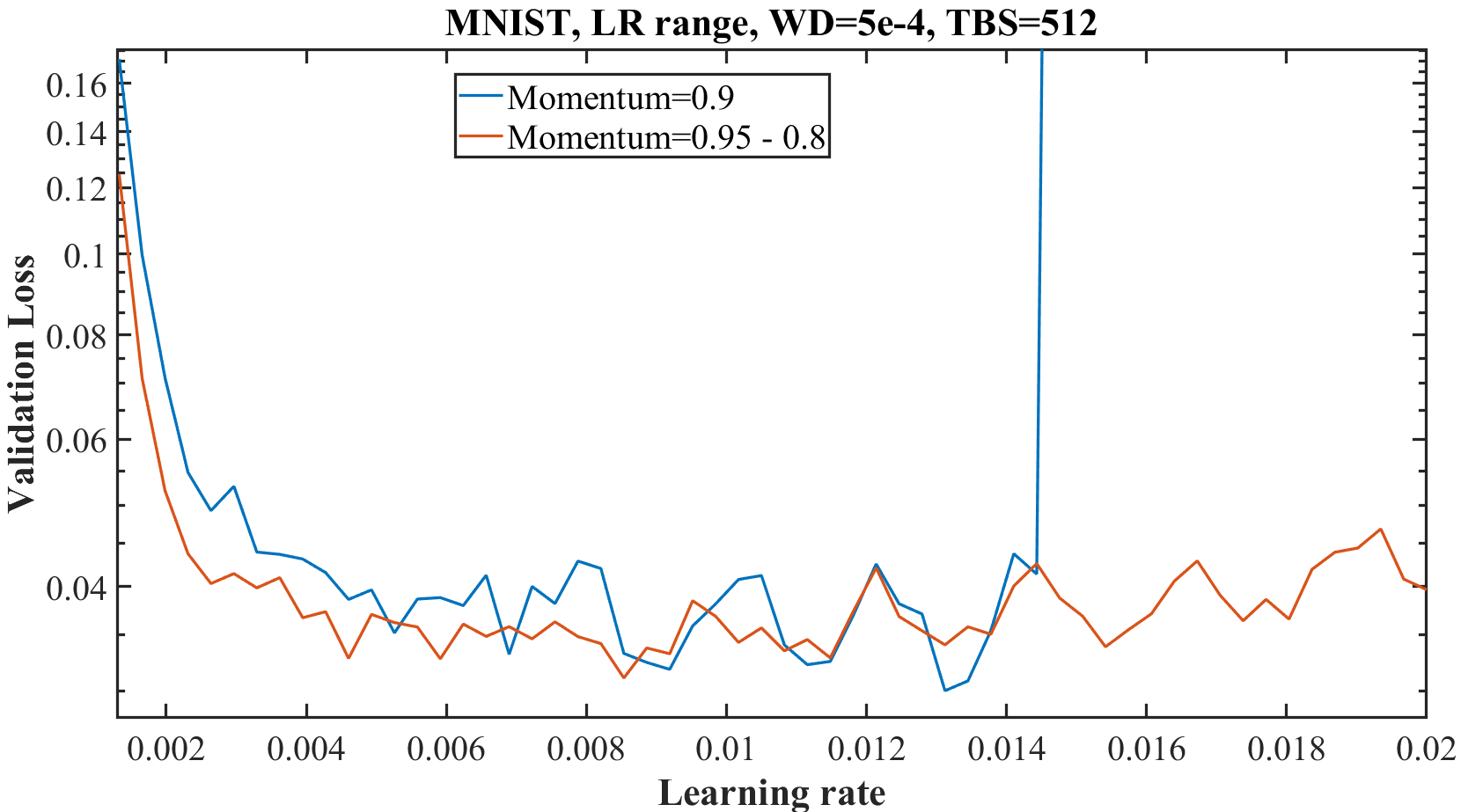}
		\caption{Comparison of momentum.}
		\label{fig:mnist3layerCM2}       
	\end{subfigure}
	\quad
	\hfill
	~ 
	\centering
	\begin{subfigure}[b]{0.46\textwidth}
		\includegraphics[width=\textwidth]{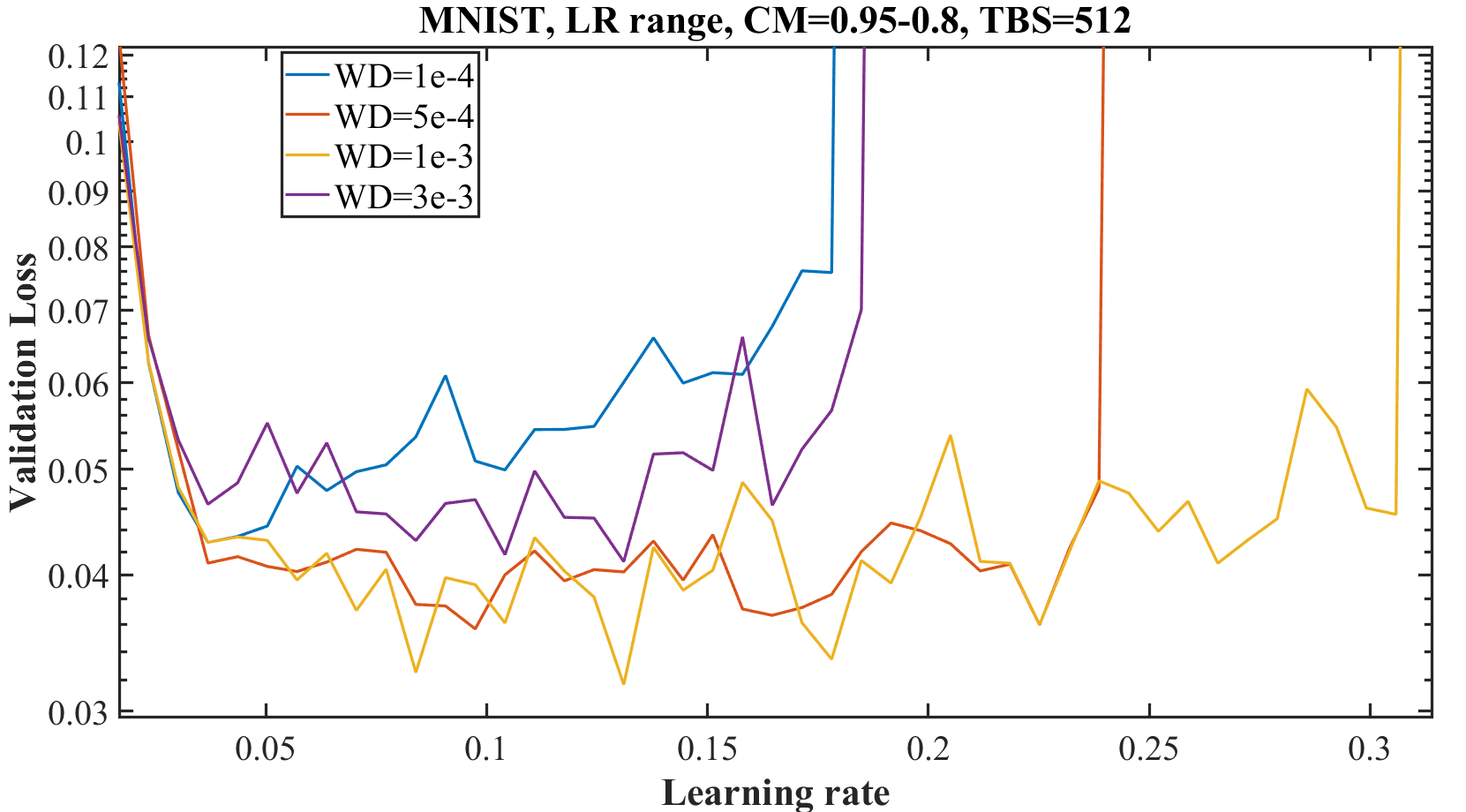}
		\caption{Comparison of weight decay.}
		\label{fig:mnist3layerWD2}       
	\end{subfigure}
	\caption{Hyper-parameter search for MNIST dataset with a shallow, 3-layer network.}
	\label{fig:mnist3layer}
	\vspace{-15pt}	
\end{figure}

\subsection{MNIST}

The MNIST database of handwritten digits from 0 to 9 (i.e., 10 classes)  has a training set of 60,000 examples, and a test set of 10,000 examples.  It is simpler than Cifar-10, so the shallow, 3-layer LeNet architecture was used for the tests.  Caffe \url{https://github.com/BVLC/caffe} provides the LeNet architecture and its associated hyper-parameters with the download of the Caffe framework.  The provided values of the hyper-parameters are; learning rate = 0.01, momentum = 0.9, and weight decay = 0.0005.  These initial values are useful for comparison.

Figure \ref{fig:mnist3layerCM2} shows a comparison of momentum during a learning rate range test from 0.001 to 0.04. The momentum default value of 0.9 is visibly inferior to a cyclical momentum decreasing from 0.95 to 0.8.  This Figure illustrates that the cyclical momentum case is flatter and permits a longer learning rate range than a constant value of 0.9.  A search for the optimal weight decay value is illustrated in Figure \ref{fig:mnist3layerWD2}.  The weight decay default value of 0.0005 provides good performance with similar result as a larger value of 0.001, which defines the optimal weight decay range.  Also note that making the weight decay both larger (i.e., 0.003) or smaller (i.e., 0.0001) increases the loss.

Table \ref{tab:otherExamples} lists the results of training the MNIST dataset with the LeNet architecture.  Using the hyper-parameters provided with the Caffe download, an accuracy of 99.03\% is obtained in 85 epochs.  Switching from an inv learning rate policy to a step policy produces equivalent results.  However, switching to the 1cycle policy produces an accuracy near 99.3\%, even in only 12 epochs.  

\begin{figure} [tbh]
	\centering
	\begin{subfigure}[b]{0.47\textwidth}
		\includegraphics[width=\textwidth]{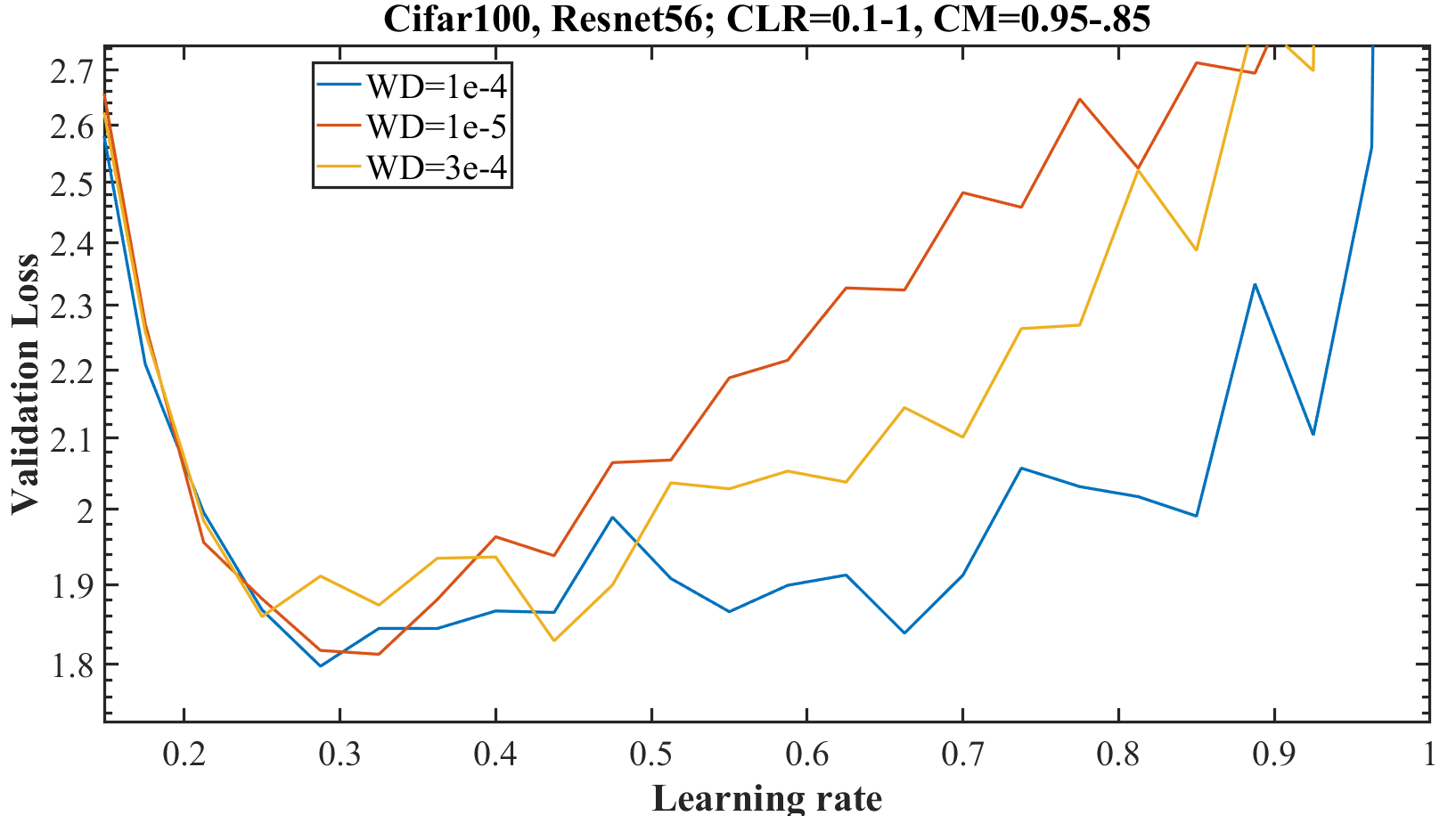}
		\caption{Weight decay search}
		\label{fig:resnet56Cifar100WD}       
	\end{subfigure}
	\quad
	\hfill
	~ 
	\centering
	\begin{subfigure}[b]{0.46\textwidth}
		\includegraphics[width=\textwidth]{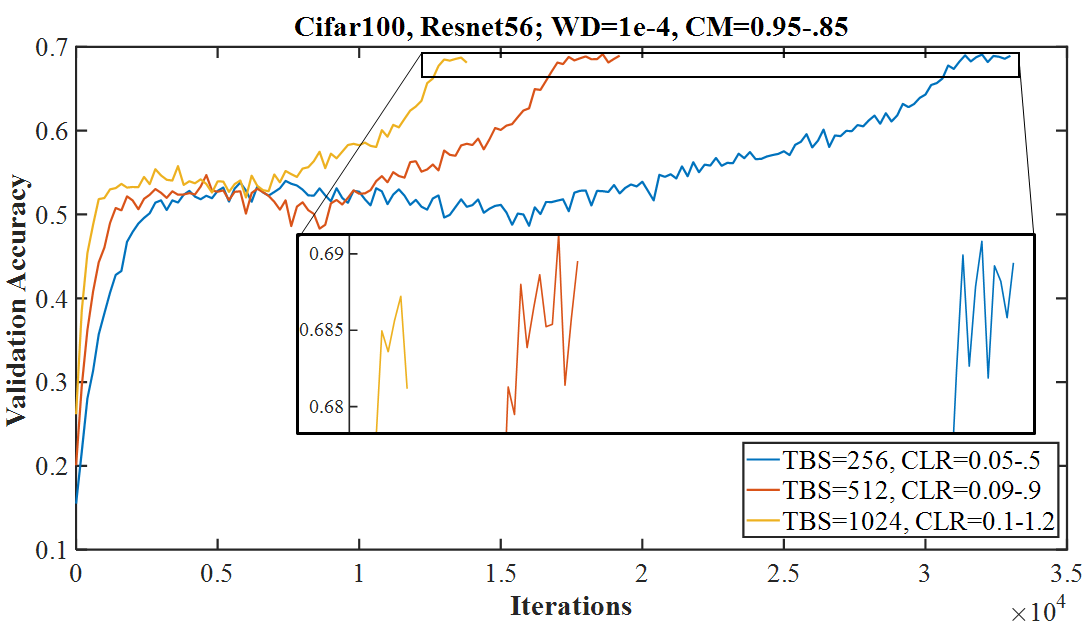}
		\caption{Total Batch Size search.}
		\label{fig:resnet56Cifar100TBS}       
	\end{subfigure}
	\caption{Hyper-parameter search for the Cifar-100 dataset with the resnet-56 architecture.}
	\label{fig:resnet56Cifar100}
	\vspace{-5pt}	
\end{figure}

\subsection{Cifar-100}
\label{sec:cifar100}

This dataset is like the CIFAR-10, except it has 100 classes. Hence, instead of 5,000 training images per class, there are only 500 training images and 100 testing images per class.   Due to the similarity to Cifar-10, one can expect the same optimal hyper-parameter values for Cifar-100 and it is verified below.  


These experiments used the same resnet-56 architecture as used for Cifar-10.  For Cifar-10 the optimal hyper-parameters are; learning rate range = 0.1 - 1, batch size = 512,  cyclical momentum = 0.95 - 0.85, and weight decay = $10^{-4}$.  Figure  \ref{fig:resnet56Cifar100WD} shows the validation loss for weight decay at values of  $ 3 \times 10^{-4}, 10^{-4} $ and $10^{-5} $.  The best value for  weight decay is $10^{-4}$ as both a larger and smaller value results in a higher loss.  

Figure \ref{fig:resnet56Cifar100TBS} compares the accuracies training curves for three batch sizes, 256, 512, 1024.  The number of training iterations/epochs were adjusted to provide similar execution times.  In this case, the accuracies are all within the standard deviations of each other.

Table \ref{tab:otherExamples} compares the final accuracies of training with a step learning rate policy to training with a 1cycle learning rate policy.  The training results from the 1cycle learning rate policy are significantly higher than than the results from the step learning rate policy.  In addition, the number of epochs required for training is reduced by an order of magnitude (i.e., even at only 25 epochs, the accuracy is higher for 1cycle than 800 epochs with a step learning rate policy).

\begin{figure} [tbh]
	\centering
	\begin{subfigure}[b]{0.47\textwidth}
		\includegraphics[width=\textwidth]{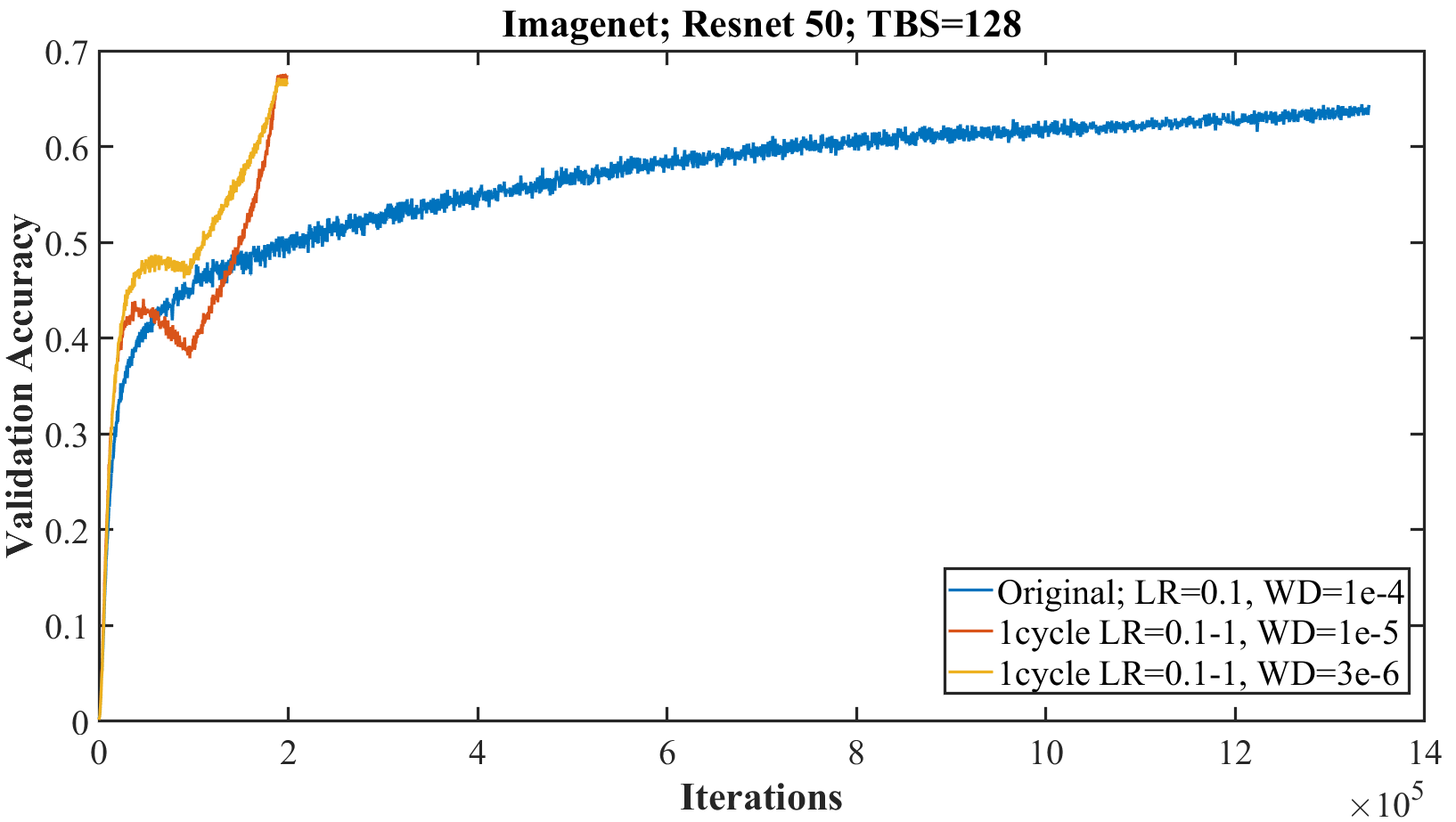}
		\caption{Resnet-50}
		\label{fig:imagenetResnetSC}       
	\end{subfigure}
	\quad
	\hfill
	~ 
	\centering
	\begin{subfigure}[b]{0.46\textwidth}
		\includegraphics[width=\textwidth]{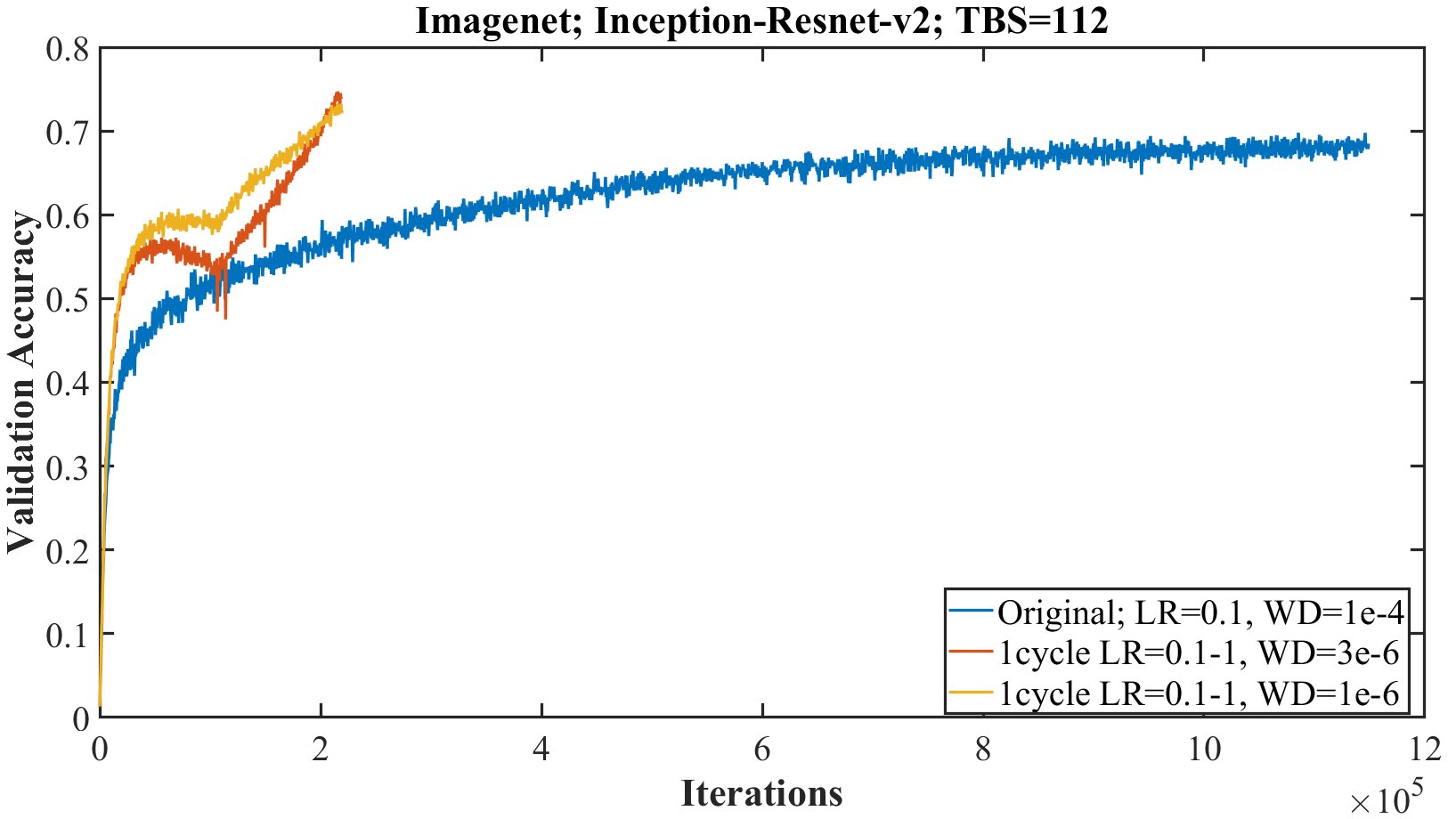}
		\caption{Inception-resnet-v2}
		\label{fig:imagenetInceptionSC}       
	\end{subfigure}
	\caption{Training resnet and inception architectures on the imagenet dataset with the standard learning rate policy (blue curve) versus a 1cycle policy that displays super-convergence.  Illustrates that deep neural networks can be trained much faster (20 versus 100 epochs) than by using the standard training methods.}
	\label{fig:imagenetResnet}
	\vspace{-5pt}	
\end{figure}


\subsection{Imagenet}
\label{sec:imagenet}

Imagenet, which is a large image database with 1.24 million training images for 1,000 classes, was downloaded from \url{http://image-net.org/download-images}.  The resnet-50 and inception-resnet-v2 architectures used for these experiments were obtained from \url{https://github.com/soeaver/caffe-model}.  
This Section shows that for high dimensional data such as Imagenet, our experiments\footnote{Due to the high computational requirements for training Imagenet, results reported here are the average of only two runs.} show that reducing or eliminating regularization in the form of weight decay allows the use of larger learning rates and produces much faster convergence and higher final accuracies.  

Figure \ref{fig:imagenetResnetSC} presents the comparison of training a resnet-50 architecture on Imagenet with the current standard training methodology versus the super-convergence method.  The hyper-parameters choices for the original training is set to the recommended values in Szegedy \etal \cite{szegedy2017inception} (i.e., momentum = 0.9, LR = 0.045 decaying every 2 epochs using an exponential rate of 0.94, WD = $10^{-4}$).  This produced the familiar training curve indicated by the blue line in the Figure.    Due to time limitations, the runs were not executed to completion of the training but the accuracy after 130 epochs is 63.7\% and the curve can be extrapolated to an accuracy near 65\%.

The red and yellow training curves in Figure \ref{fig:imagenetResnetSC} use the 1cycle learning rate schedule for 20 epochs, with the learning rate varying from 0.05 to 1.0, then down to 0.00005.  In order to use such large learning rates, it was necessary to reduce the value for weight decay.  Our tests found that weight decay values in the range from $3 \times 10^{-6}$ to $10^{-5}$ provided the best accuracy of $67.6\%$. Smaller values of weight decay showed signs of underfitting, which hurt performance  It is noteworthy that the two curves (especially for a weight decay value of $3 \times 10^{-6}$) display small amounts of overfitting.  Empirically this implies that small amounts of overfitting is a good indicator of the best value and helps the search for the optimal weight decay values early in the training.

Figure \ref{fig:imagenetInceptionSC} presents the training a inception-resnet-v2 architecture on Imagenet.  The blue curve is the current standard way while the red and yellow curves use the 1cycle learning rate policy.  The inception architecture tells a similar story as the resnet architecture.  Using the same step learning rate policy, momentum, and weight decay as in Szegedy \etal \cite{szegedy2017inception} produces the familiar blue curve in Figure \ref{fig:imagenetInceptionSC}.  After 100 epochs the accuracy is 67.6\% and the curve can be extrapolated to an accuracy in the range of 69-70\%.

On the other hand, reducing the weight decay permitted use of the 1cycle learning rate policy with the learning rate varying from 0.05 to 1.0, then down to 0.00005 in 20 epochs.  The weight decay values in the range from $3 \times 10^{-6}$ to $10^{-6}$ works well and  using a weight decay of $3 \times 10^{-6}$ provides the best accuracy of $ 74.0\%$.  As with resnet-50, there appears to be a small amount of overfitting with this weight decay value. 

The lesson from these experiments is that deep neural networks can be trained much faster than by the standard training methods.  The list at the beginning of Section \ref{sec:other} shows the way.



\section{Discussion}

This report is a step towards a more comprehensive view of training deep neural networks.  Scattered throughout the deep learning literature are papers that examine a couple of factors in isolation and propose recommendations.  Unfortunately, there are contradictory recommendations of this kind in literature, which is reminiscent of the parable of the blind men and the elephant. Our future work for Part 2 of this report includes expanding the current study to include a better understanding of the effects of data, data augmentation, network depth/width, and other forms of regularization (i.e., dropout ratio, stochastic depth).

Highlighted throughout this report are various remarks and suggestions to enable faster training of networks to produce optimal results.  The purpose is to provide instructions and make the application of neural networks easier.  New deep learning applications are continuously appearing and are obtaining significant improvements in new fields of study.  This expansion is showing no signs of slowing and it is likely to touch all of us in new ways during the next few years.

\subsubsection*{Acknowledgments}

The author expresses his appreciation of the funding from the Naval Research Laboratory base program that enabled this work. The views, positions and conclusions expressed herein reflect only the authors' opinions and expressly do not reflect those of the Office of Naval Research, nor those of the Naval Research Laboratory.

\bibliographystyle{SimpleGuideSkipConn}
\bibliography{SimpleGuideSkipConn}

\begin{thebibliography}{23}
\providecommand{\natexlab}[1]{#1}
\providecommand{\url}[1]{\texttt{#1}}
\expandafter\ifx\csname urlstyle\endcsname\relax
  \providecommand{\doi}[1]{doi: #1}\else
  \providecommand{\doi}{doi: \begingroup \urlstyle{rm}\Url}\fi

\bibitem[Aarts \& Korst(1988)Aarts and Korst]{aarts1988simulated}
Emile Aarts and Jan Korst.
\newblock Simulated annealing and boltzmann machines.
\newblock 1988.

\bibitem[Bengio(2012)]{bengio2012practical}
Yoshua Bengio.
\newblock Practical recommendations for gradient-based training of deep
  architectures.
\newblock In \emph{Neural networks: Tricks of the trade}, pp.\  437--478.
  Springer, 2012.

\bibitem[Bengio et~al.(2009)Bengio, Louradour, Collobert, and
  Weston]{bengio2009curriculum}
Yoshua Bengio, Jerome Louradour, Ronan Collobert, and Jason Weston.
\newblock Curriculum learning.
\newblock In \emph{Proceedings of the 26th annual international conference on
  machine learning}, pp.\  41--48. ACM, 2009.

\bibitem[Bergstra \& Bengio(2012)Bergstra and Bengio]{bergstra2012random}
James Bergstra and Yoshua Bengio.
\newblock Random search for hyper-parameter optimization.
\newblock \emph{Journal of Machine Learning Research}, 13\penalty0
  (Feb):\penalty0 281--305, 2012.

\bibitem[Goodfellow et~al.(2016)Goodfellow, Bengio, Courville, and
  Bengio]{goodfellow2016deep}
Ian Goodfellow, Yoshua Bengio, Aaron Courville, and Yoshua Bengio.
\newblock \emph{Deep learning}, volume~1.
\newblock MIT press Cambridge, 2016.

\bibitem[He et~al.(2016)He, Zhang, Ren, and Sun]{he2016deep}
Kaiming He, Xiangyu Zhang, Shaoqing Ren, and Jian Sun.
\newblock Deep residual learning for image recognition.
\newblock In \emph{Proceedings of the IEEE conference on computer vision and
  pattern recognition}, pp.\  770--778, 2016.

\bibitem[Hern{\'a}ndez-Garc{\'\i}a \& K{\"o}nig(2018)Hern{\'a}ndez-Garc{\'\i}a
  and K{\"o}nig]{hernandez2018deep}
Alex Hern{\'a}ndez-Garc{\'\i}a and Peter K{\"o}nig.
\newblock Do deep nets really need weight decay and dropout?
\newblock \emph{arXiv preprint arXiv:1802.07042}, 2018.

\bibitem[Jastrzebski et~al.(2017a)Jastrzebski, Arpit, Ballas, Verma, Che, and
  Bengio]{jastrzebski2017residual}
Stanis{\l}aw Jastrzebski, Devansh Arpit, Nicolas Ballas, Vikas Verma, Tong Che,
  and Yoshua Bengio.
\newblock Residual connections encourage iterative inference.
\newblock \emph{arXiv preprint arXiv:1710.04773}, 2017a.

\bibitem[Jastrzebski et~al.(2017b)Jastrzebski, Kenton, Arpit, Ballas, Fischer,
  Bengio, and Storkey]{jastrzkebski2017three}
Stanis{\l}aw Jastrzebski, Zachary Kenton, Devansh Arpit, Nicolas Ballas, Asja
  Fischer, Yoshua Bengio, and Amos Storkey.
\newblock Three factors influencing minima in sgd.
\newblock \emph{arXiv preprint arXiv:1711.04623}, 2017b.

\bibitem[Kingma \& Ba(2014)Kingma and Ba]{kingma2014adam}
Diederik Kingma and Jimmy Ba.
\newblock Adam: A method for stochastic optimization.
\newblock \emph{arXiv preprint arXiv:1412.6980}, 2014.

\bibitem[Kuka{\v{c}}ka et~al.(2017)Kuka{\v{c}}ka, Golkov, and
  Cremers]{kukavcka2017regularization}
Jan Kuka{\v{c}}ka, Vladimir Golkov, and Daniel Cremers.
\newblock Regularization for deep learning: A taxonomy.
\newblock \emph{arXiv preprint arXiv:1710.10686}, 2017.

\bibitem[Liu et~al.(2018)Liu, Chen, Zhou, and Zhao]{liu2018toward}
Tianyi Liu, Zhehui Chen, Enlu Zhou, and Tuo Zhao.
\newblock Toward deeper understanding of nonconvex stochastic optimization with
  momentum using diffusion approximations.
\newblock \emph{arXiv preprint arXiv:1802.05155}, 2018.

\bibitem[Lorraine \& Duvenaud(2018)Lorraine and
  Duvenaud]{lorraine2018stochastic}
Jonathan Lorraine and David Duvenaud.
\newblock Stochastic hyperparameter optimization through hypernetworks.
\newblock \emph{arXiv preprint arXiv:1802.09419}, 2018.

\bibitem[Orr \& M{\"u}ller(2003)Orr and M{\"u}ller]{orr2003neural}
Genevieve~B Orr and Klaus-Robert M{\"u}ller.
\newblock \emph{Neural networks: tricks of the trade}.
\newblock Springer, 2003.

\bibitem[Smith(2015)]{smith2015no}
Leslie~N Smith.
\newblock No more pesky learning rate guessing games.
\newblock \emph{arXiv preprint arXiv:1506.01186}, 2015.

\bibitem[Smith(2017)]{smith2017cyclical}
Leslie~N Smith.
\newblock Cyclical learning rates for training neural networks.
\newblock In \emph{Applications of Computer Vision (WACV), 2017 IEEE Winter
  Conference on}, pp.\  464--472. IEEE, 2017.

\bibitem[Smith \& Topin(2017)Smith and Topin]{smith2017super}
Leslie~N Smith and Nicholay Topin.
\newblock Super-convergence: Very fast training of residual networks using
  large learning rates.
\newblock \emph{arXiv preprint arXiv:1708.07120}, 2017.

\bibitem[Smith \& Le(2017)Smith and Le]{smith2017understanding}
Samuel~L Smith and Quoc~V Le.
\newblock Understanding generalization and stochastic gradient descent.
\newblock \emph{arXiv preprint arXiv:1710.06451}, 2017.

\bibitem[Smith et~al.(2017)Smith, Kindermans, and Le]{smith2017don}
Samuel~L Smith, Pieter-Jan Kindermans, and Quoc~V Le.
\newblock Don't decay the learning rate, increase the batch size.
\newblock \emph{arXiv preprint arXiv:1711.00489}, 2017.

\bibitem[Srivastava et~al.(2014)Srivastava, Hinton, Krizhevsky, Sutskever, and
  Salakhutdinov]{srivastava2014dropout}
Nitish Srivastava, Geoffrey Hinton, Alex Krizhevsky, Ilya Sutskever, and Ruslan
  Salakhutdinov.
\newblock Dropout: A simple way to prevent neural networks from overfitting.
\newblock \emph{The Journal of Machine Learning Research}, 15\penalty0
  (1):\penalty0 1929--1958, 2014.

\bibitem[Szegedy et~al.(2017)Szegedy, Ioffe, Vanhoucke, and
  Alemi]{szegedy2017inception}
Christian Szegedy, Sergey Ioffe, Vincent Vanhoucke, and Alexander~A Alemi.
\newblock Inception-v4, inception-resnet and the impact of residual connections
  on learning.
\newblock In \emph{AAAI}, volume~4, pp.\ ~12, 2017.

\bibitem[Wilson \& Martinez(2003)Wilson and Martinez]{wilson2003general}
D~Randall Wilson and Tony~R Martinez.
\newblock The general inefficiency of batch training for gradient descent
  learning.
\newblock \emph{Neural Networks}, 16\penalty0 (10):\penalty0 1429--1451, 2003.

\bibitem[Xing et~al.(2018)Xing, Arpit, Tsirigotis, and Bengio]{xing2018walk}
Chen Xing, Devansh Arpit, Christos Tsirigotis, and Yoshua Bengio.
\newblock A walk with sgd.
\newblock \emph{arXiv preprint arXiv:1802.08770}, 2018.

\end{thebibliography}

\appendix


\section{Appendix}

\subsection{Experimental methods: detailed information about the experiments to enable replication}
\label{sec:exp}

The hardware primarily used for these experiments was an IBM Power8, 32 compute nodes, 20 cores/node, 4 Tesla P100 GPUs/node with 255 GB available memory/node.  An additional server that was used for some experiments was a 64 node cluster with 8 Nvidia Titan Black GPUs, 128 GB memory, and dual Intel Xenon E5-2620 v2 CPUs per node.  

To improve the generality of these results, each curve in the plots is an average of four runs, each with slightly different batch sizes and different initialization schemes.  This was done to marginalize the batch size and initialization to help insure that the results are independent of precise choices.  Two even and two odd batch sizes were used within in a range from a minimum batch size to that size plus 12.  Total batch size reported is an average batch size of the four runs.  The two initialization schemes were msra and xavier and the runs alternated between these two.




\subsection{Implementation of cyclical momentum in Caffe}

In the file SGDSolver, and the function ComputeUpdateValue, replace the line:
 \begin{lstlisting}
 Dtype momentum = this->param_.momentum();
\end{lstlisting}
(near line 314) with:
\begin{lstlisting}
Dtype momentum = this->param_.momentum();
if (this->param_.cyclical_momentum_size() == 2) {
  int cycle = this->iter_  /
    (2*this->param_.cyclical_momentum(1));
  float x = (float) (this->iter_ - 
    (2*cycle+1)*this->param_.cyclical_momentum(1));
  x = x / this->param_.cyclical_momentum(1);
  momentum  = this->param_.momentum() + 
    (this->param_.cyclical_momentum(0)- this->param_.momentum()) *
    std::min(double(1), std::max(double(0), 
    (1.0 - fabs(x))/pow(2.0,double(cycle))));
}
\end{lstlisting}
This code assumes the addition of a new solver input variable ``cyclical\_momentum'' that needs to be added to caffe.proto.  Hence, in the file caffe.proto, in message SolverParameter add near line 174:

\begin{lstlisting}
repeated float cyclical_momentum = 44; 
\end{lstlisting}
This enables reading in the cyclical\_momentum variable twice in your solver.prototxt file.  An example use of cyclical momentum in the solver file is:
\begin{lstlisting}
momentum: 0.9
cyclical_momentum: 0.95
cyclical_momentum: 10000
\end{lstlisting}
where the training will start with a momentum of 0.9 and linearly increase the momentum to 0.95 by iteration 10,000, then decrease the momentum to 0.9 by iteration 20,000.  This cycle will continue for the duration of the training.

\end{document}